\documentclass[11pt,a4paper]{article}
\usepackage[hyperref]{emnlp2018}
\usepackage{times}
\usepackage{latexsym}
\usepackage{color}

\usepackage{url}

\aclfinalcopy % Uncomment this line for the final submission
  
\usepackage{amsmath}
\usepackage{amssymb}
\usepackage{enumitem}
\usepackage{mathtools}
\usepackage{multirow}
\usepackage{tcolorbox}
\newcommand{\denselist}{\setlength{\itemsep}{1pt}
  \setlength{\parskip}{0pt} \setlength{\parsep}{0pt}}
\newcommand{\bitem}{\begin{itemize}[noitemsep,topsep=2pt]\denselist}
\newcommand{\eitem}{\end{itemize}}
\newcommand{\qed}{\square}

\newfont{\msym}{msbm10}
\newcommand{\reals}{\mbox{\msym R}}

\usepackage{bm}

\DeclareMathOperator{\E}{\mathbb{E}}
\newcommand{\var}{\operatorname{Var}}
\newcommand{\mse}{\operatorname{MSE}}
\newcommand{\tr}{\operatorname{Tr}}
\newcommand{\wt}[1]{\widetilde{#1}}
\newcommand{\wh}[1]{\widehat{#1}}
\newcommand{\gtheta}{\ensuremath \nabla_{\theta}}

\newcommand{\mle}{\wh{\theta}^{\text{~MLE}}}

\newcommand{\objr}{{\cal L}_R^n}
\newcommand{\objri}{{\cal L}_R^{\infty}}
\newcommand{\objb}{{\cal L}_B^n}

\newcommand{\gbeta}{\nabla_{\beta}}

\newcommand{\px}{p_X}

\newcommand{\q}[1]{q(#1|x, y_{0:K}; \theta^*)}

\newcommand{\str}[3]{s(#1, #2; #3)}
\newcommand{\ssf}[3]{\bar{s}(#1, #2; #3)}
\newcommand{\sbc}[3]{\wt{s}(#1, #2; #3)}

\newcommand{\vbk}{\wt{V}_{K}}
\newcommand{\vb}{\wt{V}}
\newcommand{\wbk}{\wt{W}_{K}}
\newcommand{\wb}{\wt{W}}

\newcommand{\mub}{\wt{\mu}}
\newcommand{\mubxk}{\wt{\mu}_{x, K}}
\newcommand{\mubx}{\wt{\mu}_{x}}

\newcommand{\vnorm}[1]{\left\| #1 \right\|}

\newcommand{\fisher}{\mathcal{I}_{\theta^*}}

\newtheorem{theorem}{Theorem}[section]

\newtheorem{assumption}{Assumption}[section]
\newtheorem{lemma}{Lemma}[section]

\newtheorem{remark}{Remark}[section]

\DeclareMathOperator*{\argmax}{argmax}

\newcommand{\commentout}[1]{}
\newcommand{\loss}{\hbox{objective}}

\title{Noise Contrastive Estimation and Negative Sampling
for Conditional Models: Consistency and Statistical Efficiency}

\author{Zhuang Ma \\
  University of Pennsylvania\thanks{$\;\,$Part of this work done at Google.}\\
  {\tt zhuangma@wharton.upenn.edu} \\\And
  Michael Collins \\
  Google AI Language and Columbia University\thanks{$\;\,$Work done at Google.} \\
  {\tt mjcollins@google.com} \\}

\date{}

\begin{document}
\maketitle
\begin{abstract}
Noise Contrastive Estimation (NCE) is a powerful parameter estimation
method for log-linear models, which avoids calculation of the
partition function or its derivatives at each training step, a
computationally demanding step in many cases. It is closely related
to {\em negative sampling methods}, now widely used in NLP.
This paper considers
NCE-based estimation of {\em conditional} models. Conditional models
are frequently encountered in practice; however there has not been a
rigorous theoretical analysis of NCE in this setting, and we will
argue there are subtle but important questions when generalizing NCE
to the conditional case. In particular, we analyze two variants of NCE
for conditional models: one based on a classification $\loss$, the other
based on a ranking $\loss$. We show that the ranking-based variant of NCE
gives consistent parameter estimates under weaker assumptions than the
classification-based method; we analyze the statistical efficiency of
the ranking-based and classification-based variants of NCE; finally we
describe experiments on synthetic data and language modeling showing the effectiveness
and trade-offs of both methods.
%that the ranking-based method outperforms the classification-based method in practice.
\end{abstract}

\newcommand{\sss}{s}

\section{Introduction}
This paper considers parameter estimation in conditional models of
the form
\begin{equation}
 p(y | x; \theta) = \frac{\exp \left ( \sss(x, y; \theta) \right )}{Z(x; \theta)}
%p(y | x; \theta) = \exp \left ( \sss(x, y; \theta) \right )/ Z(x; \theta)
\label{eq:first}
\end{equation}
where $\sss(x, y; \theta)$ is the unnormalized score of label $y$ in
conjunction with input $x$ under parameters $\theta$, ${\cal Y}$ is
a finite set of possible labels, and $Z(x; \theta) = \sum_{y \in
  {\cal Y}} \exp \left ( \sss(x, y; \theta) \right )$ is the partition
function for input $x$ under parameters $\theta$.

It is hard to overstate the importance of models of this form in NLP.
In log-linear models, including both the original work on
maximum-entropy models \cite{maxent}, and later work on conditional
random fields \cite{crfs}, the scoring function $\sss(x, y; \theta) =
\theta \cdot f(x, y)$ where $f(x, y) \in \reals^d$ is a feature
vector, and $\theta \in \reals^d$ are the parameters of the model.  In
more recent work on neural networks the function $\sss(x, y; \theta)$
is a non-linear function. In Word2Vec the
scoring function is
$
\sss(x, y; \theta) = \theta_x \cdot \theta'_y
$
where $y$ is a word in the context of word $x$, and $\theta_x \in \reals^d$
and $\theta'_y \in \reals^d$ are ``inside'' and ``outside'' word embeddings 
$x$ and $y$.

In many NLP applications the set ${\cal Y}$ is large.
\commentout{: for example in
language modeling, Word2Vec learning, or machine translation using
seq2seq models, the set ${\cal Y}$ is a vocabulary that can easily
have 10s or even 100s of thousands of words.}
Maximum likelihood
estimation (MLE) of the parameters $\theta$ requires calculation of
$Z(x; \theta)$ or its derivatives at each training step, thereby
requiring a summation over all members of ${\cal Y}$, which can be
computationally expensive. 
This has led to many authors considering alternative methods,
often
referred to as ``negative sampling methods'', where a
modified training objective is used that does not require summation over
${\cal Y}$ on each example. Instead negative examples are drawn
from some distribution, and a $\loss$ function is derived based on binary
classification or ranking. Prominent examples are the binary objective
used in word2vec (\cite{Mikolov:2013}, see also
\cite{Levy:2014:NWE}), and the Noise Contrastive Estimation methods of
\cite{mnih2012fast,jozefowicz2016exploring} for estimation of language
models.

In spite of the centrality of negative sampling methods, they are
arguably not well understood from a theoretical standpoint. There are
clear connections to noise contrastive estimation
(NCE) \cite{gutmann2012noise}, a negative sampling method for
parameter estimation in {\em joint} models of the form
\begin{equation}
p(y ) = \frac{\exp \left ( \sss(y; \theta) \right )}{Z(\theta)};\;\;\;
Z(\theta) = \sum_{y \in {\cal Y}} \exp \left ( \sss(y; \theta) \right )
\label{eq:second}
\end{equation}
However there has not been a rigorous theoretical analysis of NCE in
the estimation of {\em conditional} models of the form in
Eq.~\ref{eq:first}, and we will argue there are subtle but important
questions when generalizing NCE to the conditional case. 
In particular, the joint model in Eq~\ref{eq:second} has a single
partition function $Z(\theta)$ which is estimated as a parameter
of the model \cite{gutmann2012noise} whereas the conditional model
in Eq~\ref{eq:first} has a separate partition function $Z(x; \theta)$
for each value of $x$. This difference is critical.

We show the following (throughout we define $K \geq 1$ to
be the number of negative examples sampled per
training example):

%In this paper, we analyze two variants of NCE for conditional models:
%one based on binary classification loss, the other based on ranking
%loss. We include the following results:

% {\color{blue} There are subtle but important questions when generalizing NCE to the conditional case: each conditional distribution $p(y | x; \theta)$ has its own partition function $Z(x; \theta)$. Without further assumptions, the classification-based NCE does not guarantee consistency. We remark that when ${\cal X}$ is a singleton, the conditional model is reduced to the model in the original NCE setup. }

$\bullet$ For any $K \geq 1$, a binary classification variant of NCE, as
  used by \cite{mnih2012fast,Mikolov:2013}, gives consistent parameter estimates under the
  assumption that $Z(x; \theta)$ is constant with respect to $x$
  (i.e., $Z(x; \theta) = H(\theta)$ for some function
  $H$). Equivalently, the method is consistent under the assumption
  that the function $\sss(x, y; \theta)$ is powerful enough to
  incorporate $\log Z(x; \theta)$.

$\bullet$ For any $K \geq 1$, a ranking-based variant of NCE, as used by
  \cite{jozefowicz2016exploring}, gives consistent parameter estimates under the much
  weaker assumption that $Z(x; \theta)$ can vary with
  $x$. Equivalently, there is no need for $\sss(x, y; \theta)$ to be
  powerful enough to incorporate $\log Z(x; \theta)$.

$\bullet$ We analyze the statistical efficiency of the ranking-based and
  classification-based NCE variants. Under respective assumptions,
  both variants achieve Fisher efficiency (the same asymptotic
  mean square error as the MLE) as $K \rightarrow \infty$.

$\bullet$ We discuss application of our results to approaches of
\cite{mnih2012fast,Mikolov:2013,Levy:2014:NWE,jozefowicz2016exploring}
giving a unified account of these 
methods.

$\bullet$ We describe experiments on synthetic data and language modeling evaluating the
effectiveness of the two NCE variants.
%ranking-based vs. binary-classification methods.

\vspace{-1.5ex}
\section{Basic Assumptions}

We assume the following setup throughout:
%Throughout this paper we assume the following setup:

$\bullet$ We have sets ${\cal X}$ and ${\cal Y}$, where ${\cal X}, {\cal Y}$ are
  finite. 

$\bullet$ There is some unknown joint distribution $p_{X, Y}(x, y)$ where
  $x \in {\cal X}$ and $y \in {\cal Y}$. We assume that the marginal distributions satisfy $p_X(x) >0$ for all $x \in {\cal X}$ and
  $p_Y(y)> 0$ for all $y\in {\cal Y}$. 

$\bullet$ We have training examples $\{x^{(i)}, y^{(i)}\}_{i=1}^n$
drawn I.I.D. from $p_{X, Y}(x, y)$.

 $\bullet$ We have a scoring function $\str{x}{y}{\theta}$ where $\theta$
   are the parameters of the model. For example, $\str{x}{y}{\theta}$ may
   be defined by a neural network.

$\bullet$ We use $\Theta$ to refer to the parameter space.
%(set of possible  values of $\theta$). 
We assume that $\Theta \subseteq \reals^d$ for some integer $d$.

$\bullet$ We use $p_N(y)$ to refer to a distribution from which
negative examples are drawn in the NCE approach. We assume that
$p_N$ satisfies $p_N(y) > 0$ for all $y\in {\cal Y}$.

\commentout{
\begin{itemize}

\item We have sets ${\cal X}$ and ${\cal Y}$, where ${\cal X}, {\cal Y}$ are
  finite. 

\item There is some unknown joint distribution $p_{X, Y}(x, y)$ where
  $x \in {\cal X}$ and $y \in {\cal Y}$. We assume that the marginal distributions satisfy $p_X(x) >0$ for all $x \in {\cal X}$ and
  $p_Y(y)> 0$ for all $y\in {\cal Y}$. 

\item We have training examples $\{x^{(i)}, y^{(i)}\}_{i=1}^n$
drawn I.I.D. from $p_{X, Y}(x, y)$.

 \item We have some scoring function $\str{x}{y}{\theta}$ where $\theta$
   are the parameters of the model. For example, $\str{x}{y}{\theta}$ may
   be defined by a neural network.

\item We use $\Theta$ to refer to the parameter space (set of possible
  values of $\theta$). We assume that $\Theta \subseteq \reals^d$ for some integer $d$.

\item We use $p_N(y)$ to refer to a distribution from which
negative examples are drawn in the NCE approach. We assume that
$p_N(y) > 0$ for all $y\in {\cal Y}$.

\end{itemize}}

We will consider estimation under the following two assumptions:

\begin{assumption}
There exists some parameter value $\theta^* \in \Theta$ such that for all $ (x, y)\in\mathcal{X} \times \mathcal{Y}$, 
\begin{equation}
 p_{Y|X}(y | x) = \frac{\exp ( \str{x}{y}{\theta^*} ) }{Z(x; \theta^*)}
%   p_{Y|X}(y | x) = \exp ( \str{x}{y}{\theta^*} ) / Z(x; \theta^*)
  \label{eq:model-ranking}
\end{equation}
where $Z(x; \theta^*) = \sum_{y \in {\cal Y}} \exp ( \str{x}{y}{\theta^*} )
$.
\label{assump:ranking}
\end{assumption}
\begin{assumption}
There exists some parameter value $\theta^* \in \Theta$, and a
constant $\gamma^* \in \reals$, such that for all $ (x, y)\in\mathcal{X} \times \mathcal{Y}$, 
\begin{equation}
  p_{Y|X}(y | x) = \exp \left( \str{x}{y}{\theta^*} - \gamma^* \right).
  \label{eq:model-binary}
\end{equation}
\label{assump:binary}
\end{assumption}

\vspace{-2.5ex}

Assumption~\ref{assump:binary} is stronger than
Assumption~\ref{assump:ranking}. It requires $\log Z(x; \theta^*) \equiv \gamma^*$ for all $x\in{\cal X}$, that is, the conditional distribution is perfectly self-normalized. 
%Notice that $p_{Y|X}(y | x)$ is probability density for any $x\in{\cal X}$. So 
%under 
Under Assumption~\ref{assump:binary}, it must be the case that $\forall x\in {\cal X}$
\begin{equation*}
  \sum_y p_{Y|X}(y | x) = \sum_y \exp\{\str{x}{y}{\theta^*} - \gamma^*\}
 = 1
 % \label{eq:model-binary}
\end{equation*}
There are $|{\cal X}|$ constraints but only $d+1$ free parameters. 
Therefore self-normalization is a non-trivial assumption when $|{\cal X}| \gg d$.
In the case of language modeling, $|{\cal X}| = |V|^k \gg d+1$, where $|V|$ is the vocabulary size and $k$ is 
the length of the context. The number of constraints grows exponentially fast. 
\commentout{MJC:commented this out for now at least.
Finally we remark that, by taking expectation over $x$ on both sides of the equlity above, $\gamma^*$ is uniquely determined by $\theta^*$: 
\[
  \exp(\gamma^*) = \sum_x\sum_y p_X(x)\exp\{\str{x}{y}{\theta^*}\}.
\]
}

Given a scoring function $s(x, y; \theta)$ that satisfies
assumption~\ref{assump:ranking}, we can derive a scoring function
$s'$ that satisfies assumption~\ref{assump:binary} by defining
\[
s'(x, y; \theta, \{c_x: x \in {\cal X}\}) = s(x, y; \theta) - c_x
\]
where $c_x \in \reals$ is a parameter for history $x$. Thus we
introduce a new parameter $c_x$ for each possible history $x$. This is
the most straightforward extension of NCE to the conditional
case; it is used by \cite{mnih2012fast}. It has the clear
drawback however of introducing a large number of additional
parameters to the model.

\section{Two Estimation Algorithms}

\commentout{
This section first introduces the two negative sampling algorithms.
Section~\ref{sec:overview} then give an informal overview of consistency results for the two
algorithms (section~\ref{??} of this paper gives the full proof: we
include the informal discussion to give more intuition). Finally 
section~\ref{related}
discusses the use of the two algorithms in previous work.}

Figure~\ref{estimation1} shows two NCE-based parameter estimation
algorithms, based respectively on {\em binary $\loss$} and {\em ranking
  $\loss$}. The input to either algorithm is a set of training
examples $\{x^{(i)}, y^{(i)}\}_{i=1}^n$, a parameter $K$ specifying
the number of negative examples per training example, and a
distribution $p_N(\cdot)$ from which negative examples are
sampled. The algorithms differ only in the choice of $\loss$ function
being optimized: $L^n_B$ for binary $\loss$, and $L^n_R$ for ranking
$\loss$. Binary $\loss$ essentially corresponds to a problem where the
scoring function $\str{x}{y}{\theta}$ is used to construct a binary
classifier that discriminates between
positive and negative examples. Ranking $\loss$ corresponds to a problem
where the scoring function $\str{x}{y}{\theta}$ is used to rank the
true label $y^{(i)}$ above negative examples $y^{(i, 1)} \ldots y^{(i, K)}$
for the input $x^{(i)}$. 

Our main result is as follows:
%The following informal theorem states our main result:

\begin{theorem}(Informal: see section~\ref{sec:theory} for a formal statement.)
% Assume the negative sampling distribution satisfies $p_N(y) > 0$ for any $y$ such that $p_Y(y) > 0$ where $p_Y(\cdot)$ is the marginal distribution of $Y$. 
For any $K \geq 1$, the binary classification-based algorithm in
figure~\ref{estimation1} is consistent under
Assumption~\ref{assump:binary}, but is not always consistent under the
weaker Assumption~\ref{assump:ranking}. For any $K \geq 1$, the
ranking-based algorithm in figure~\ref{estimation1} is consistent under
either Assumption~\ref{assump:ranking} or
Assumption~\ref{assump:binary}. Both algorithms achieve the same
statistical efficiency as the maximum-likelihood estimate as $K \rightarrow
\infty$.
\end{theorem}

The remainder of this section gives a sketch of the argument
underlying consistency, and discusses use of the two
algorithms in previous work.

\begin{figure}[t!]
\framebox{\parbox{\columnwidth}{
\begin{small}
%\begin{tcolorbox}[colback=white]
% \begin{tcolorbox}[colback=white, width=0.9\textwidth]
{\bf Inputs:} Training examples $\{x^{(i)}, y^{(i)}\}_{i=1}^n$,
sampling distribution $p_N(\cdot)$ for generating negative examples,
an integer $K$ specifying the number of negative examples per training
example, a scoring function $\str{x}{y}{\theta}$.  Flags $\{
\hbox{BINARY = true, RANKING = false} \}$ if binary classification
objective is used, $\{ \hbox{BINARY = false, RANKING = true} \}$ if ranking
objective is used.
\medskip

{\bf Definitions:} Define $\ssf{x}{y}{\theta} = \str{x}{y}{\theta} - \log p_N(y)$
%\\ hence $\exp(\ssf{x}{y}{\theta}) = \exp(\str{x}{y}{\theta})/p_N(y)$.

{\bf Algorithm:}

\begin{itemize}

\item For $i = 1 \ldots n$, $k = 1 \ldots K$, draw $y^{(i, k)}$
  I.I.D. from the distribution $p_{N}(y)$. For
convenience define $y^{(i,0)} = y^{(i)}$.

\item If RANKING, define the ranking objective function
\[
\objr (\theta) = \frac{1}{n}\sum_{i=1}^n \log \frac{ \exp ( \ssf{x^{(i)}}{y^{(i, 0)}}{\theta} ) 
}
{\sum_{k=0}^K \exp (
  \ssf{x^{(i)}}{y^{(i, k)}}{\theta} ) },
%{\exp ( \ssf{x^{(i)}}{y^{(i)}}{\theta} ) + \sum_{k=1}^K \exp (
%  \ssf{x^{(i)}}{y^{(i, k)}}{\theta} ) },
\]
and the estimator $\wh{\theta}_R  =  \argmax\limits_{\theta \in \Theta} \objr(\theta).$

\item If BINARY, define the binary objective function
\begin{equation*}
\begin{aligned}
\;\hspace{-0.7cm}
\objb(\theta, \gamma)  =& \frac{1}{n}\sum_{i=1}^n \left\{  \log g(x^{(i)}, y^{(i, 0)}; \theta, \gamma) \right .\\ 
& \hspace{0.7cm} 
+  \sum_{k=1}^K
\left . \log \left(1 - g(x^{(i)}, y^{(i, k)}; \theta, \gamma)\right)\right\}, 
\end{aligned}
\end{equation*}
and estimator $(\wh{\theta}_B, \wh{\gamma}_B) = \argmax\limits_{\theta \in \Theta, \gamma \in
  \Gamma} \objb(\theta, \gamma),$
where
\[
g(x, y; \theta, \gamma) = \frac{\exp \left(\ssf{x}{y}{\theta} - \gamma\right) }{\exp
  \left(\ssf{x}{y}{\theta} - \gamma\right) + K}.
\]

\item Define $\wh{\theta} = \wh{\theta}_R$ if RANKING and $\wh{\theta} = \wh{\theta}_B$ otherwise. Return $\wh{\theta}$ and 
\[
\wh{p}_{Y|X}(y | x) = 
\frac{\exp ( \str{x}{y}{\wh{\theta}} ) }{\sum_{y \in {\cal Y}} \exp  (\str{x}{y}{\wh{\theta}} )}
\]
\end{itemize}
\end{small}
}}
%\end{tcolorbox}
\caption{Two NCE-based estimation algorithms, using ranking objective and binary objective respectively.}
\label{estimation1}
\end{figure}

%!TEX root = emnlp2018_all.tex

\subsection{A Sketch of the Consistency Argument for the Ranking-Based Algorithm}
\label{sec:overview}

\newcommand{\ce}{\hbox{C}}

\newcommand{\bary}{\bar{y}}

In this section, in order to develop intuition underlying the ranking
algorithm, we give a proof sketch of the following theorem:
\begin{theorem} (First part of theorem~\ref{thm:pop-consistency-rank} below.)
Define $\objri(\theta) = \E[ \objr(\theta)]$.
Under Assumption~\ref{assump:ranking}, $\bar{\theta}\in \arg\max_\theta \objri(\theta)$ if and only if, for all $ (x, y)\in\mathcal{X} \times \mathcal{Y}$,
$$
p_{Y|X}(y|x) = \exp(\str{x}{y}{\bar{\theta}})/Z(x, \bar{\theta} ).
$$
\label{th:fp}
\end{theorem}
\vspace{-0.5cm}
This theorem is key to the consistency argument.
Intuitively as $n$ increases $\objr(\theta)$ converges to
$\objri(\theta)$, and the output to the algorithm converges to
$\theta'$ such that $p(y | x; \theta') = p_{Y|X}(y | x)$ for all $x,
y$. Section~\ref{sec:theory} gives a formal argument.

\commentout{
The input to the algorithm is a set $\{x^{(i)},
y^{(i)}\}$ for $i = 1 \ldots n$ drawn from the distribution $p(x, y)$.
In addition for each training example we sample $K$ ``negative''
examples $y^{(i,k)}$ for $k = 1 \ldots K$.}

We now give a proof sketch for theorem~\ref{th:fp}.
Consider the algorithm in figure~\ref{estimation1}.
For convenience define
$\bar{y}^{(i)}$ to be the vector $(y^{(i,0)}, y^{(i,1)}, \ldots, y^{(i, K)})$.
Define 
$\alpha(x, \bary) = \sum_{k=0}^K p_{X,Y}(x, \bary_k) \prod_{j \neq k} p_N(\bary_j)$,
and
\begin{eqnarray*}
 q(k | x, \bary; \theta) &=&
\frac{ \exp ( \ssf{x}{\bary_k}{\theta} ) }
{
\sum_{k=0}^K \exp (
  \ssf{x}{\bary_k}{\theta} ) },\\
\beta(k | x, \bary) &=&
\frac{p_{X, Y}(x, \bary_k) \prod_{j \neq k} p_N(\bary_j)}
{\alpha(x, \bary)}
\\
&=& \frac{p_{Y|X}(\bary_k| x)/p_N(\bary_k)}{\sum_{k=0}^N p_{Y|X}(\bary_k|x)/p_N(\bary_k)}\\
\hbox{$\;$\hspace{-1.7cm}$\;$}
\ce(x, \bary; \theta) = 
&&
\hbox{$\;$\hspace{-1cm}$\;$}
- \sum_{k=0}^K \beta(k | x, \bary) \log q(k | x, \bary; \theta)
\end{eqnarray*}
Intuitively, $q(\cdot|x, \bary; \theta)$ and $\beta(\cdot|x, \bary)$ are posterior
distributions over the true label $k \in \{0 \ldots K\}$ given an input $x, \bary$,
under the parameters $\theta$ and the true distributions $p_{X, \bar{Y}}(x, \bary)$
respectively; $\ce(x, \bary; \theta)$ is the negative cross-entropy
between these two distributions.

\commentout{
Thus $q(\ldots; \theta)$ is a distribution under parameters $\theta$,
$\beta(\ldots)$ is a distribution under the true underlying distribution,
and $\ce(\ldots)$ is the negative cross entropy between these two
distributions.}

\commentout{
The proof of theorem~\ref{th:fp} rests on two identities.
Note that $\objr(\theta) = \frac{1}{n} \sum_{i=1}^n \log q(0 | x^{(i)}, \bary^{(i)}; \theta)$.
The first identity is as follows:
\begin{small}
\begin{eqnarray}
&&\hbox{$\;$\hspace{-1cm}$\;$} \objri(\theta) = \E[L^n_R(\theta)] \nonumber\\
& \hbox{$\;$\hspace{-1cm}$\;$}=& \sum_x \sum_{\bary} p(x, \bary_0) \prod_{k=1}^K p_N(\bary_k) \log q(0 | x, \bary; \theta) \label{eq:ei1}\\
& \hbox{$\;$\hspace{-1cm}$\;$}=& \frac{1}{K + 1} \sum_x \sum_{\bary} \alpha(x, \bary) \ce(x, \bary; \theta)
\label{eq:hr}
\end{eqnarray}
\end{small}
Eq.~\ref{eq:ei1} follows because
each pair $x^{(i)}, \bary^{(i)}$ is drawn from the distribution
$
p_{X,\bar{Y}}(x, \bary) = p_{X,Y}(x, \bary_0) \prod_{k=1}^K p_N(\bary_k)
$. Eq.~\ref{eq:hr} follows through algebraic manipulations.
}

The proof of theorem~\ref{th:fp} rests on two identities. The first identity states that the objective function is the expectation of the negative cross-entropy w.r.t. the density function $\frac{1}{K + 1} \alpha(x, \bary)$ (see Section~\ref{sec:pf-pop-consis-ranking} of the supplementary material for derivation):
\begin{eqnarray}
 \objri(\theta) =  \sum_x \sum_{\bary} \frac{1}{K + 1} \alpha(x, \bary) \ce(x, \bary; \theta).
\label{eq:hr}
\end{eqnarray}
The second identity concerns the relationship between $q(\cdot |x, \bary; \theta)$
and $\beta(\cdot | x, \bary)$.  Under assumption~\ref{assump:ranking}, for all $x, \bary$, $k \in \{0 \ldots K\}$,
\begin{eqnarray}
&&q(k | x, \bary; \theta^*) \nonumber \\
&=& \frac{p_{Y|X}(\bary_k | x) Z(x; \theta^*)/{p_N(y_k)}}{\sum_{k=0}^K
    p_{Y|X}(\bary_k | x) Z(x; \theta^*)/{p_N(y_k)}} \nonumber \\
&=& \beta(k | x, \bary) \label{eq:beta}
\end{eqnarray}
It follows immediately through the properties of negative cross entropy that
\begin{equation}
\forall x, \bary, \;\;
\theta^* \in \argmax_\theta \ce(x, \bary; \theta)
\label{eq:ts}
\end{equation}

The remainder of the argument is as follows:

$\bullet$ Eqs.~\ref{eq:ts} and~\ref{eq:hr} imply that $\theta^* \in \argmax_\theta \objri(\theta)$.

$\bullet$ Assumption~\ref{assump:ranking} implies that $\alpha(x, \bary) >
  0$ for all $x, \bary$. It follows that any $\theta' \in \arg\max_\theta \objri(\theta)$
satisfies
\begin{eqnarray}
\hbox{for all $x, \bary, k, \hspace{4cm}$} && \label{eq:qconstraints}
\\
q(k | x, \bary; \theta') = q(k | x, \bary;
\theta^*) = \beta(k | x, \bary) \nonumber
\end{eqnarray}
Otherwise there would be some $x, \bary$ such that $\ce(x, \bary; \theta') < \ce(x, \bary; \theta^*)$.

$\bullet$ Eq.~\ref{eq:qconstraints} implies that $\forall x, y$, $p(y
  | x; \theta') = p(y | x; \theta^*)$. See the proof of lemma~\ref{lem:iden}
  in the supplementary material.

In summary, the identity in Eq.~\ref{eq:hr} is key: the objective
function in the limit, $\objri(\theta)$, is
related to a negative cross-entropy between the
underlying distribution $\beta(\cdot|x, \bary)$ and a distribution
under the parameters, $q(\cdot|x, \bary; \theta)$. The parameters $\theta^*$
maximize this negative cross entropy over the space of all
distributions $\{q(\cdot|x, \bary; \theta), \theta \in \Theta \}$.

\subsection{The Algorithms in Previous Work}

\label{related}

To motivate the importance of the two algorithms, we
now discuss their application in previous work.

\citet{mnih2012fast} consider language modeling, where $x = w_1 w_2
\ldots w_{n-1}$ is a history consisting of the previous $n-1$ words,
and $y$ is a word. The scoring function is defined as
\[
s(x, y; \theta) = (\sum_{i=1}^{n-1} C_i r_{w_i}) \cdot q_y + b_y - c_x
\]
where $r_{w_i}$ is an embedding (vector of parameters) for history
word $w_i$, $q_y$ is an embedding (vector of parameters) for word $y$,
each $C_i$ for $i = 1 \ldots n-1$ is a matrix of parameters specifying
the contribution of $r_{w_i}$ to the history representation, $b_y$ is
a bias term for word $y$, and $c_x$ is a parameter corresponding to
the log normalization term for history $x$. Thus each history $x$ has
its own parameter $c_x$. The binary $\loss$ function is used in the NCE
algorithm. The noise distribution $p_N(y)$ is set to be the unigram
distribution over words in the vocabulary.

This method is a direct application of the original NCE method
to conditional estimation, through introduction of the parameters
$c_x$ corresponding to normalization terms for each history. Interestingly,
\citet{mnih2012fast} acknowledge the difficulties in maintaining a
separate parameter $c_x$ for each history, and set $c_x = 0$ for all $x$,
noting that empirically this works well, but without giving justification.

\citet{Mikolov:2013} 
consider an NCE-based method using the binary
$\loss$ function for estimation of word embeddings. The skip-gram method
described in the paper corresponds to a model where $x$ is a word, and
$y$ is a word in the context. The vector $v_x$ is the embedding for
word $x$, and the vector $v'_y$ is an embedding for word $y$ (separate
embeddings are used for $x$ and $y$). The method they describe uses 
\[
\ssf{x}{y}{\theta} = v'_y \cdot v_x
\]
or equivalently
\[
\str{x}{y}{\theta} = v'_y \cdot v_x + \log p_N(y)
\]
The negative-sampling distribution $p_N(y)$ was chosen as the unigram distribution
$p_Y(y)$ raised to the power $3/4$. The end goal of the method was to learn
useful embeddings $v_w$ and $v'_w$ for each word in the vocabulary; 
\commentout{however assuming that Assumption~\ref{assump:binary} holds, the method gives a consistent
estimate for a model of the form
\begin{eqnarray*}
p(y | x) &=& \frac{\exp \left ( v'_y \cdot v_x + \log p_N(y) \right )}{\sum_y \exp \left ( v'_y \cdot v_x + \log p_N(y) \right )}\\
&=&\frac{p_N(y) \exp \left ( v'_y \cdot v_x \right )}{Z(x; \theta)}
\end{eqnarray*}
where $Z(x; \theta) = \sum_y p_N(y) \exp \left ( v'_y \cdot v_x \right
) = H(\theta)$ does not vary with $x$. }
however the method gives a consistent
estimate for a model of the form
\begin{eqnarray*}
p(y | x) &=& \frac{\exp \left ( v'_y \cdot v_x + \log p_N(y) \right )}{\sum_y \exp \left ( v'_y \cdot v_x + \log p_N(y) \right )}\\
&=&\frac{p_N(y) \exp \left ( v'_y \cdot v_x \right )}{Z(x; \theta)}
\end{eqnarray*}
assuming that Assumption~\ref{assump:binary} holds, i.e. $Z(x; \theta) = \sum_y p_N(y) \exp \left ( v'_y \cdot v_x \right
) \equiv H(\theta)$ which does not vary with $x$.

\citet{Levy:2014:NWE} make a connection between the NCE-based method
of \cite{Mikolov:2013}, and factorization of a matrix of
pointwise mutual information (PMI) values of $(x, y)$ pairs. 
Consistency of the NCE-based
method under assumption~\ref{assump:binary} implies a similar result,
specifically: if we define $p_N(y) = p_Y(y)$, and define
$\str{x}{y}{\theta} = v'_y \cdot v_x + \log p_N(y)$
implying $\ssf{x}{y}{\theta} = v'_y \cdot v_x$,
then parameters
$v'_y$ and $v_x$ converge to values such that
\begin{eqnarray*}
p(y | x) &=& 
\frac{p_Y(y) \exp \left ( v'_y \cdot v_x \right )}{H(\theta)}
\end{eqnarray*}
or equivalently
\begin{eqnarray*}
\hbox{PMI}(x, y) &=& \log \frac{p(y | x)}{p(y)} = 
v'_y \cdot v_x - \log H(\theta)
\end{eqnarray*}
That is, following \cite{Levy:2014:NWE}, the
inner product $v'_y \cdot v_x$ is an estimate of the PMI
up to a constant offset $H(\theta)$.

Finally, \citet{jozefowicz2016exploring} introduce the ranking-based
variant of NCE for the language modeling problem. This is 
the same as the ranking-based algorithm in figure~\ref{estimation1}.
They do not, however, make the connection to
assumptions~\ref{assump:binary} and~\ref{assump:ranking}, or derive
the consistency or efficiency results in the current paper.
\citet{jozefowicz2016exploring} partially motivate the ranking-based
variant throught the {\em importance sampling} viewpoint of
\citet{bengio2008adaptive}. However there are two critical 
differences: 1) the algorithm of \citet{bengio2008adaptive} does
not lead to the same objective $L^n_R$ in the ranking-based
variant of NCE; instead it uses importance sampling to
derive an objective that is similar but not identical; 2) 
the importance sampling method leads to a biased estimate
of the gradients of the log-likelihood function, with the bias
going to zero only as $K \rightarrow \infty$. In contrast the
theorems in the current paper show that the NCE-based methods
are {\em consistent for any value of $K$}. In summary, while
it is tempting to view the ranking variant of NCE as an importance
sampling method, the NCE-based view gives stronger guarantees for
finite values of $K$.

\section{Theory}
\label{sec:theory}

This section states the main theorems. The supplementary material
contains proofs. Throughout the paper, we use $\E_{X}[\,\cdot\,], \E_{Y}[\,\cdot\,], \E_{X, Y}[\,\cdot\,], \E_{Y|X=x}[\,\cdot\,]$
to represent the expectation w.r.t. $p_X(\cdot), p_Y(\cdot)$, $p_{X, Y}(\cdot, \cdot),p_{Y|X}(\cdot|x)$. 
% This section gives the main theorems. The supplementary material
% contains proofs. Throughout the paper, we use 
% \[
% \E_{X}[\,\cdot\,], \E_{Y}[\,\cdot\,], \E_{X, Y}[\,\cdot\,], \E_{Y|X=x}[\,\cdot\,]
% \]
% to represent the expectation w.r.t. 
% \[
% p_X(\cdot), p_Y(\cdot), p_{X, Y}(\cdot, \cdot),p_{Y|X}(\cdot|x)
% \]
% respectively and reserve $\E[\,\cdot\,]$ for the
% expectation over the joint distribution of all random variables.
% The same convention is adopted for the variance operator
% $\var [\,\cdot\,]$. 
We use $\|\cdot\|$ to denote either the $l_2$
norm when the operand is a vector or the spectral norm when the
operand is a matrix. Finally, we use $\Rightarrow$ to represent converge in distribution.
% We use $\|\cdot\|$ to denote either the $l_2$
% norm of a vector or the spectral norm of a matrix.
Recall that we have defined 
\[
\ssf{x}{y}{\theta} = \str{x}{y}{\theta} - \log p_N(y).
\]

\commentout{
In practice, $p_Y(\cdot)$ is
unknown but can be substituted by its sample estimate, e.g. the
empirical distribution of $y$. 
\commentout{
This paper focuses on the regime where
the sample size is large enough, to the extent that the conditional
distribution $p_{Y|X}(\cdot\,|\,\cdot)$ can be well estimated, that is
$n >> |\mathcal{X}\times \mathcal{Y}|$. Under this condition, the
estimation of $p_Y(\cdot)$ is a second order issue compared with the
estimation of the scoring function $s(x, y;\theta)$. }
Technically, to
avoid this simplification, one can divide the training samples into
two parts, using the first part to construct an estimator of
$p_Y(\cdot)$, denoted by $\wh{p}_Y(\cdot)$, and run the noise
contrastive estimation algorithm on the second part with
$\wt{p}_Y(\cdot) = \wh{p}_Y(\cdot)$. This guarantees the negative
samples $(y^{(i, 1)}, \cdots, y^{(i, K)})$ drawn from
$\wt{p}_Y(\cdot)$ and the training example $(x^{(i)}, y^{(i)})$ are
independent such that the asymptotic analysis in the proof will go
through.}

\subsection{Ranking}
\label{sec:ranking}
In this section, we study noise contrastive estimation with ranking $\loss$ under Assumption~\ref{assump:ranking}. 
% We first justify the rationale of optimizing the ranking objective $\objr(\theta)$ on the population level. 
First consider the following function:
%Under certain moment conditions, as $n\rightarrow\infty$, by law of
%large numbers, the ranking objective function $\objr(\theta)$
%converges to
\begin{equation*}
\begin{aligned}
L_R^\infty(\theta) & = \sum_{x, y_0, \cdots, y_K} p_{X, Y}(x, y_0)\prod_{i=1}^K p_N(y_i)  \\
& \times  \log\left(\frac{\exp(\ssf{x}{y_0}{\theta})}{\sum_{k=0}^K \exp( \ssf{x}{y_k}{\theta})}\right).  
\end{aligned}
\end{equation*}
%For any value of $\theta$ such that $L_R^\infty(\theta) < \infty$, we
%have
%\[
%{\cal P} \{
%\lim_{n \rightarrow \infty} L_R^n(\theta) = L_R^\infty(\theta) \} = 1
%\]
By straightforward calculation, one can find that 
\[
L_R^\infty(\theta) = \E \left[
  L_R^n(\theta)\right].
\]
Under mild conditions, $L^n_R(\theta)$
converges to $L^\infty_R(\theta)$ as $n \rightarrow \infty$. 
% ; we will shortly make this rigorous, but we first analyze the properties of
% $L^\infty_R(\theta)$.
Denote the set of maximizers of $L^\infty_R(\theta)$ by $\Theta_R^* $, that is 
 \[
 \Theta_R^* = \arg\max_{\theta\in\Theta} L_R^\infty\left(\theta\right).
 \]
% $\Theta_R^* = \left\{\bar{\theta}~ |~ \bar{\theta} \in \arg\max_{\theta\in\Theta} L_R^\infty\left(\theta\right)\right\}.$
The following theorem shows that any parameter vector $\bar{\theta}
\in \Theta_R^*$ if and only if it gives the correct conditional distribution $p_{Y|X}(y
| x)$. 
% It also gives a necessary and sufficient condition for the model parameters $\theta^*$ to be identifiable.

% justifies the rationale of optimizing the
% ranking objective $\objr(\theta)$. It also shows that
% Assumption~\ref{assump:identifiability-rank} is necessary and
% sufficient for the parameter $\theta^*$ in Assumption~\ref{assump:ranking} to
% be identifiable.
\begin{assumption}
  (Identifiability). 
For any $\theta\in\Theta$, if there exists a function
$c(x)$ such that $\str{x}{y}{\theta} - \str{x}{y}{\theta^*} \equiv c(x)$
for all $ (x, y)\in\mathcal{X} \times \mathcal{Y}$, then $\theta = \theta^*$ and thus 
$c(x) = 0$ for all $x$.
  \label{assump:identifiability-rank}
\end{assumption}

\begin{theorem} 
Under Assumption~\ref{assump:ranking}, $\bar{\theta}\in\Theta_R^*$ if and only if, for all $ (x, y)\in\mathcal{X} \times \mathcal{Y}$,
\[
p_{Y|X}(y|x) = \exp(\str{x}{y}{\bar{\theta}})/Z(x, \bar{\theta} ).
\]
In addition, $\Theta_R^*$ is a singleton if and only if Assumption~\ref{assump:identifiability-rank} holds. 
\label{thm:pop-consistency-rank}
\end{theorem}

% {\em Proof.} See Section~\ref{sec:pf-pop-consis-ranking}. 

Next we consider consistency of the estimation algorithm based on the 
ranking $\loss$ under the following regularity assumptions:
% Further, with mild assumptions on the scoring function, the estimate obtained from optimizing the finite-sample objective $\objr(\theta)$ is consistent. 
\begin{assumption}
  (Continuity). $\str{x}{y}{\theta}$ is continuous w.r.t. $\theta$ for all $(x, y)\in\mathcal{X}\times \mathcal{Y}$.
  \label{assump:continuity}
\end{assumption}
\begin{assumption} 
$\Theta_R^*$ is contained in the interior of a compact set $\Theta\subset\reals^d$.
  \label{assump:compactness-rank}
\end{assumption}

For a given estimate $\wh{p}_{Y|X}$ of the conditional distribution $p_{Y|X}$, define the error metric $d(\cdot, \cdot)$ by 
\begin{equation*}
\begin{aligned}
 d \left(\wh{p}_{Y|X}, p_{Y|X}\right) & = \sum_{x\in\mathcal{X}, y\in\mathcal{Y}} p_{X, Y}(x, y)
\\
& \times \left(\wh{p}_{Y|X}(y|x) - p_{Y|X}(y|x)\right)^2. 
\end{aligned}
\end{equation*}
% \[
% d \left(\wh{p}_{Y|X}, p_{Y|X}\right) = \sum_{x\in\mathcal{X}, y\in\mathcal{Y}} p_{X, Y}(x, y)\left(\wh{p}_{Y|X}(y|x) - p_{Y|X}(y|x)\right)^2. 
% \]
% For a sequence of IID observations $(x^{(1)}, y^{(1)})$,
% $(x^{(2)}, y^{(2)}), \ldots, $ define the sequences of estimates
% $\wh{\theta}_R^1, \wh{\theta}_R^2, \ldots$ and $\wh{p}^1_{Y|X}(y|x),
% \wh{p}^2_{Y|X}(y|x), \ldots$ where $\wh{p}^n_{Y|X}(y|x)$ is obtained by optimizing the  ranking $\loss$ on $(x^{(1)}, y^{(1)}), (x^{(2)}, y^{(2)}),
% \ldots, (x^{(n)}, y^{(n)})$.
For a sequence of IID observations $(x^{(1)}, y^{(1)})$, $(x^{(2)}, y^{(2)}), \ldots, $ define the sequences of estimates
$(\wh{\theta}_R^1, ~\wh{p}^1_{Y|X}), ~(\wh{\theta}_R^2, ~\wh{p}^2_{Y|X}), \ldots$ where the $n^{th}$ estimate $(\wh{\theta}_R^n, ~\wh{p}^n_{Y|X})$ is obtained by optimizing the ranking objective of figure~\ref{estimation1} on $(x^{(1)}, y^{(1)}), (x^{(2)}, y^{(2)}),
\ldots, (x^{(n)}, y^{(n)})$.
\begin{theorem}
  (Consistency) Under Assumptions~\ref{assump:ranking}, \ref{assump:continuity}, \ref{assump:compactness-rank}, the
  estimates based on the ranking objective are strongly consistent in the
  sense that for any fixed $K \geq 1$,
  \begin{equation*}
  \begin{aligned}
 & \mathbb{P} \Big\{\lim_{n\rightarrow \infty} \min_{\theta^*\in\Theta^*_R}\| \wh{\theta}^n_R - \theta^*\| = 0 \Big\}  \\
& = \mathbb{P} \Big\{\lim_{n\rightarrow \infty} d \left(\wh{p}^n_{Y|X},~ p_{Y|X}\right) = 0  \Big\}  = 1 
  \end{aligned}
  \end{equation*}
Further, if Assumption~\ref{assump:identifiability-rank} holds, 
\[
\mathbb{P} \left\{\lim\limits_{n\rightarrow \infty} \wh{\theta}^n_R = \theta^* \right\} = 1.
\]
\label{thm:consistency-rank}
\end{theorem}

\begin{remark}
Thoughout the paper, all NCE estimators are defined for some fixed $K$. We suppress the dependence on $K$ to simplify notation (e.g. $\wh{\theta}^n_R$ should be interpreted as $\wh{\theta}^{n, K}_{R}$).
\end{remark}

\subsection{Classification}
\label{sec:binary}
Now we turn to the analysis of NCE with binary $\loss$ under Assumption~\ref{assump:binary}. First consider the following function, 
\begin{equation*}
\begin{aligned}
L_B^\infty & \left(\theta, \gamma \right) = \sum_{x, y} \Big\{ p_{X, Y}(x, y)\log\left( g(x, y; \theta, \gamma) \right) \\
&  + K p_X(x)p_N(y) \log\left( 1 - g(x, y; \theta, \gamma) \right) \Big\}
\label{eq:binary-inf}
\end{aligned}
\end{equation*}
One can find that 
 \[
 L_B^\infty(\theta, \gamma) = \E \left[ L_B^n(\theta,
  \gamma)\right].
  \]
%   Under mild conditions, 
% $L^n_B(\theta, \gamma)$ converges to $L^\infty_B(\theta, \gamma)$ as
% $n \rightarrow \infty$.  
Denote the set of maximizers of
$L^\infty_B(\theta, \gamma)$ by $\Omega_B^*:$
\[ 
\Omega_B^* = \arg\max_{\theta\in\Theta,
  \gamma\in\Gamma} L_B^\infty\left(\theta, \gamma\right).
  \]
% $\Omega_B^* = \left\{(\bar{\theta}, \bar{\gamma}) ~|~ (\bar{\theta}, \bar{\gamma}) \in \arg\max_{\theta\in\Theta, \gamma\in\Gamma} L_B^\infty\left(\theta, \gamma\right)\right\}$. 
Parallel results of Theorem~\ref{thm:pop-consistency-rank}, \ref{thm:consistency-rank} are established as follows. 
\begin{assumption}
  (Identifiability). For any $\theta\in\Theta$, if there exists some constant
$c$ such that $\str{x}{y}{\theta} - \str{x}{y}{\theta^*} \equiv c~$ for all $ (x, y)\in\mathcal{X} \times \mathcal{Y}$, then $\theta = \theta^*$ and thus $c=0$.
  \label{assump:identifiability-binary}
\end{assumption}

\begin{assumption} 
  $\Omega_B^*$ is in the interior of $\Theta\times\Gamma$ where $\Theta\subset\reals^d, \Gamma\subset\reals$ are compact sets.
    \label{assump:compactness-binary}
\end{assumption}
\begin{theorem}
Under Assumption~\ref{assump:binary}, 
% $(\bar{\theta}, \bar{\gamma})$ maximizes $L_B^\infty\left(\theta, \gamma \right)$ 
$(\bar{\theta}, \bar{\gamma})\in\Omega_B^*$ if and only if, for all $ (x, y)\in\mathcal{X} \times \mathcal{Y}$,
\[
p_{Y|X}(y|x) = \exp(\str{x}{y}{\bar{\theta}} - \bar{\gamma}) 
\]
% \[
% p_X(x)p_Y(y)exp(\ssf{x}{y}{\bar{\theta}} + \bar{\gamma}) = p_{X, Y}(x, y)
% \]
for all $(x, y)$. $\Omega_B^*$ is a singleton if and only if
Assumption~\ref{assump:identifiability-binary} holds.
% The maximizer of $L_B^\infty\left(\theta, \gamma
% \right)$ is unique ($\Omega_B^*$ is a singleton) if and only if
% Assumption~\ref{assump:identifiability-binary} holds.
\label{thm:pop-consistency-binary}
\end{theorem}

% {\em Proof.} See Section~\ref{sec:pf-pop-consis-binary}. 

% For a sequence of IID observations $(x^{(1)}, y^{(1)})$, $(x^{(2)},
% y^{(2)}), \ldots$, define the sequences of estimates
% $(\wh{\theta}_B^1, \wh{\gamma}_B^1), (\wh{\theta}_B^2,
% \wh{\gamma}_B^2), \ldots$ and $\wh{p}^1_{Y|X}(y|x),
% \wh{p}^2_{Y|X}(y|x), \ldots$ where $\wh{p}^n_{Y|X}(y|x)$ is obtained by optimizing the  classification loss on $(x^{(1)}, y^{(1)}), (x^{(2)}, y^{(2)}),
% \ldots, (x^{(n)}, y^{(n)})$.

Similarly we can define the sequence of estimates
$(\wh{\theta}_B^1,~ \wh{\gamma}_B^1, ~\wh{p}^1_{Y|X}), ~(\wh{\theta}_B^2,~
\wh{\gamma}_B^2,~ \wh{p}^2_{Y|X}), \ldots$  based on the binary objective.

% For a sequence of IID observations $(x^{(1)}, y^{(1)})$, $(x^{(2)},
% y^{(2)}), \ldots$, define the sequence of estimates
% $(\wh{\theta}_B^1,~ \wh{\gamma}_B^1, ~\wh{p}^1_{Y|X}), ~(\wh{\theta}_B^2,~
% \wh{\gamma}_B^2,~ \wh{p}^2_{Y|X}), \ldots$ where $(\wh{\theta}_B^n,~
% \wh{\gamma}_B^n, ~\wh{p}^n_{Y|X} )$ is obtained by optimizing the classification $\loss$ on $(x^{(1)}, y^{(1)})$, $(x^{(2)}, y^{(2)}),
% \ldots, (x^{(n)}, y^{(n)})$ in figure~\ref{estimation1}.

 % on $(x^{(1)}, y^{(1)}), (x^{(2)}, y^{(2)}), \ldots, (x^{(n)}, y^{(n)})$. 
%% We can then state the following consistency theorem:

\begin{theorem}
  (Consistency) Under Assumption~\ref{assump:binary},
  \ref{assump:continuity}, \ref{assump:compactness-binary}, the
  estimates defined by the binary objective are strongly consistent
  in the sense that for any $K\geq 1$,
  \begin{equation*}
  \begin{aligned}
& \mathbb{P} \Big\{\lim_{n\rightarrow \infty} \min_{(\theta^*, \gamma^*)\in\Omega_B^*}\| (\wh{\theta}_B^n, \wh{\gamma}_B^n) - (\theta^*, \gamma^*) \| = 0 \Big\} \\
& = \mathbb{P} \Big\{\lim_{n\rightarrow \infty} d \left(\wh{p}_{Y|X}^n,~ p_{Y|X}\right) = 0  \Big\}  = 1 
  \end{aligned}
  \end{equation*}
% \[
% \mathbb{P} \Big\{\lim_{n\rightarrow \infty} \min_{(\theta^*, \gamma^*)\in\Omega_B^*}\| (\wh{\theta}_B^n, \wh{\gamma}_B^n) - (\theta^*, \gamma^*) \| = 0 \Big\} = \mathbb{P} \Big\{\lim_{n\rightarrow \infty} d \left(\wh{p}_{Y|X}^n,~ p_{Y|X}\right) = 0  \Big\}  = 1
% \]
If further Assumption~\ref{assump:identifiability-binary} holds, 
\[
\mathbb{P} \left\{\lim_{n\rightarrow \infty} (\wh{\theta}_B^n, \wh{\gamma}_B^n) = \left(\theta^*, \gamma^*\right) \right\} = 1.
\]
% \[
% \mathbb{P} \left\{\lim_{n\rightarrow \infty} (\wh{\theta}_B, \wh{\gamma}_B) = \left(\theta^*, \gamma^*\right) \right\} = 1
% \]
% $(\wh{\theta}_B, \wh{\gamma}_B)$ is also strongly consistent, that is
\label{thm:consistency-binary}
\end{theorem}

% {\em Proof.} See Section~\ref{sec:pf-consis-binary}. 

\subsection{Counterexample}
In this section, we give a simple example to demonstrate that the binary classification approach fails to be consistent when assumption~\ref{assump:ranking} holds but assumption~\ref{assump:binary} fails (i.e. the partition function depends on the input). 

Consider $X\in\mathcal{X} = \{x_1, x_2\}$ with marginal distribution 
\[
p_X(x_1) = p_X(x_2) = 1/2,
\]
and $Y\in \mathcal{Y} = \{y_1, y_2\}$ generated by the conditional model specified in assumption~\ref{assump:ranking} with  
the score function parametrized by $\theta = (\theta_1, \theta_2)$ and 
\[
s(x_1, y_1; \theta) = \log \theta_1,
\]
\vspace{-0.7cm}
\[
 s(x_1, y_2; \theta) =  s(x_2, y_1; \theta) = s(x_2, y_2; \theta) = \log \theta_2.
\]
% Consider $\mathcal{X} = \{x_1, x_2\}$, $\mathcal{Y} = \{y_1, y_2\}$ with the score function parametrized by $\theta = (\theta_1, \theta_2)$ and 
% \[
% s(x_1, y_1; \theta) =  \log \theta_1,
% \]
% \vspace{-0.7cm}
% \[
% s(x_1, y_2; \theta) = s(x_2, y_1; \theta) = s(x_2, y_2; \theta) = \log \theta_2.
% \]
% Assume the marginal distribution of $X$ is
% \[
% p_X(x_1) = p_X(x_2) = 1/2,
% \]
Assume the true parameter is $\theta^* = (\theta^*_1, \theta_2^*) = (1, 3)$. By simple calculation, 
\[
Z(\theta^*; x_1) = 4, ~Z(\theta^*; x_2) = 6,
\]
\vspace{-0.7cm}
\[
p_{X, Y}(x_1, y_1) = 1/8,  p_{X, Y}(x_1, y_2) = 3/8,
\]
\vspace{-0.7cm}
\[
p_{X, Y}(x_2, y_1) = p_{X, Y}(x_2, y_2) = 1/4.
\]
Suppose we choose the negative sampling distribution $p_N(y_1) = p_N(y_2) = 1/2$. For any $K \geq 1$, by the Law of Large Numbers, as $n$ goes to infinity, $L_B^n(\theta, \gamma)$ will converge to $L_B^\infty(\theta, \gamma)$. Substitute in the parameters above. One can show that
\begin{equation*}
\begin{aligned}
L_B^\infty(\theta, \gamma) &=  \frac{1}{8} \log \frac{2\theta_1 }{2\theta_1+ K\exp(\gamma)}  \\
& + \frac{K}{4} \log \frac{K\exp(\gamma)}{2\theta_1  + K\exp(\gamma)}  \\
& + \frac{7}{8} \log \frac{2\theta_2 }{2\theta_2  + K \exp(\gamma)} \\
& + \frac{3K}{4} \log \frac{K\exp(\gamma)}{2\theta_2  + K\exp(\gamma)} .
\end{aligned}
\end{equation*}
% By the Law of Large Numbers, as $n$ goes to infinity, $L_B^n(\theta, \gamma)$ will converge to $L_B^\infty(\theta, \gamma)$. For any negative sampling distribution $p_N(y)$ satisfying $p_N(y_1), p_N(y_2) > 0$ and any $K \geq 1$, substitute in the parameters above. One can show that
% \begin{equation*}
% \begin{aligned}
% L_B^\infty(\theta, \gamma) &=  \frac{1}{8} \log \frac{\theta_1 }{\theta_1+ Kp_N(y_1)\exp(\gamma)}  \\
% & + \frac{Kp_N(y_1)}{2} \log \frac{Kp_N(y_1)\exp(\gamma)}{\theta_1  + Kp_N(y_1)\exp(\gamma)}  \\
% & + \frac{5}{8} \log \frac{\theta_2 }{\theta_2  + Kp_N(y_2)\exp(\gamma)} \\
% & + Kp_N(y_2) \log \frac{Kp_N(y_2)\exp(\gamma)}{\theta_2  + Kp_N(y_2)\exp(\gamma)} \\
% & + \frac{1}{4} \log \frac{\theta_2 }{\theta_2  + Kp_N(y_1)\exp(\gamma)} \\
% & + \frac{Kp_N(y_1)}{2} \log \frac{Kp_N(y_1)\exp(\gamma)}{\theta_2  + Kp_N(y_1)\exp(\gamma)} .
% \end{aligned}
% \end{equation*}
Setting the derivatives w.r.t. $\theta_1, \theta_2$ to zero, one will obtain 
\[
\theta_1 = \frac{1}{4}\exp(\gamma), ~~\theta_2 = \frac{7}{12}\exp(\gamma).
\]
% \[
% \frac{3\exp(\gamma)}{\theta_1(\theta_1 + Kp_N(y_1)\exp(\gamma))} = \frac{4Kp_N(y_1)}{\theta_1 + Kp_N(y_1)\exp(\gamma)} 
% \]
% \[
% \frac{5\exp(\gamma)}{\theta_2(\theta_2 + Kp_N(y_2)\exp(\gamma))} = \frac{4Kp_N(y_2)}{\theta_2 + Kp_N(y_2)\exp(\gamma)} 
% \]
So for any $ (\wt{\theta}_1, \wt{\theta}_2, \wt{\gamma}) \in \argmax_{\theta,\gamma} L_B^\infty\left(\theta, \gamma \right)$, $(\wt{\theta}_1, \wt{\theta}_2, \wt{\gamma})$ will satisfy the equalities above. 
Then the estimated distribution $\wt{p}_{Y|X}$ will satisfy
\[
\frac{\wt{p}_{Y|X}(y_1| x_1)}{\wt{p}_{Y|X}(y_2|x_1)} = \frac{\wt{\theta}_1}{\wt{\theta}_2} = \frac{1/4}{7 / 12}= \frac{3}{7},
\]
which contradicts the fact that 
\[
\frac{p_{Y|X}(y_1|x_1)}{p_{Y|X}(y_2|x_1)} = \frac{p_{X,Y}(x_1, y_1)}{p_{X,Y}(x_1, y_2)} = \frac{1}{3}.
\]
So the binary objective does not give consistent estimation of the conditional distribution. 

% Thus as long as $p_N(y_1) \neq \frac{3}{7} p_N(y_2)$, the classification-based method is not consistent. 

\subsection{Asymptotic Normality and Statistical Efficiency}
\label{sec:stat-efficiency}
Noise Contrastive Estimation significantly reduces the computational
complexity, especially when the label space $|\mathcal{Y}|$ is
large. It is natural to ask: does such scalability come at a cost?
Classical likelihood theory tells us, under mild conditions, the
maximum likelihood estimator (MLE) has nice properties like
asymptotic normality and Fisher efficiency. More specifically, as the sample
size goes to infinity, the distribution of the MLE will converge to a
multivariate normal distribution, and the mean square error of the MLE
will achieve the Cramer-Rao lower bound \citep{ferguson1996course}.

We have shown the consistency of the NCE estimators in Theorem~\ref{thm:consistency-rank} and
Theorem~\ref{thm:consistency-binary}. In this part of the paper, we derive their
asymptotic distribution and quantify their statistical efficiency. To
this end, we restrict ourselves to the case where $\theta^*$ is identifiable (i.e. Assumptions~\ref{assump:identifiability-rank} or
\ref{assump:identifiability-binary} hold) and the scoring function $\str{x}{y}{\theta}$ satisfies the following smoothness condition:
\begin{assumption}
  (Smoothness). The scoring function $\str{x}{y}{\theta}$ is twice continuous differentiable w.r.t. $\theta$ for all $(x, y)\in\mathcal{X}\times \mathcal{Y}$. 
  \label{assump:smoothness}
\end{assumption}

We first introduce the following maximum-likelihood estimator. 
%%,which serves as the benchmark for the statistical properties of the two NCE procedures:
\begin{equation*}
\begin{aligned}
& \mle = \arg\min_{\theta}~ L_{\text{MLE}}^n(\theta) \\
& := \arg\min_{\theta}\sum_{i=1}^n \log \left(\frac{\exp(\str{x^{(i)}}{y^{(i)}}{\theta})}{\sum_{y\in\mathcal{Y}}\exp(\str{x^{(i)}}{y}{\theta})}\right). 
\end{aligned}
\end{equation*}
% As being mentioned, this maximum likelihood estimator $\mle$ is computationally intensive. But it serves as a benchamrk to measure the statistical property of the proposed NCE estimator $\wh{\theta}_R$. 
Define the matrix 
\[
\fisher = \E_{X}\left[\var_{Y|X=x}\left[\nabla_{\theta} \str{x}{y}{\theta^*}\right]\right].
% \fisher = \E_{x\sim p_X(x)}\left[\var_{y\sim p_{Y|X}(y|x)}\left[\nabla_{\theta} \ssf{x}{y}{\theta^*}\right]\right].
\]
As shown below, $\fisher$ is essentially the Fisher information matrix under the conditional model. 
\begin{theorem}
  Under Assumption~\ref{assump:ranking}, \ref{assump:identifiability-rank}, \ref{assump:compactness-rank},  and \ref{assump:smoothness}, if $\fisher$ is non-singular, as $n \rightarrow\infty$
\[
    \sqrt{n}(\mle-\theta^*)~\Rightarrow~\mathcal{N}(0, \fisher^{-1}).
\]
  \label{thm:mle-mse}
  \vspace{-0.5cm}
\end{theorem}

% {\em Proof.} See Section~\ref{sec:pf-mle-mse}. 

For any given estimator $\wh{\theta}$, define the scaled asymptotic mean square error by 
\[
\mse_{\infty} (\wh{\theta}) = \lim_{n\rightarrow\infty} \E \left[  \vnorm{ \sqrt{\frac{n}{d}}\left(\wh{\theta} - \theta^*\right) }^2\right],
\]
where $d$ is the dimension of the parameter $\theta^*$.
Theorem~\ref{thm:mle-mse} implies that, 
\[
\mse_{\infty} (\mle) = \tr(\fisher^{-1})/d. 
\]
where $\tr(\cdot)$ denotes the trace of a matrix. According to classical MLE theory \citep{ferguson1996course}, under certain regularity conditions, this is the best achievable mean square error. So the next question to answer is: can these NCE estimators approach this limit? 
% We impose the following regularity conditions. 
\begin{assumption}
There exist positive constants $c, C$ such that $\sigma_{\min}(\fisher) \geq c$ and
\begin{equation*}
\begin{aligned}
\max_{(x, y)\in\mathcal{X}\times \mathcal{Y}} \Big\{ |\str{x}{y}{\theta^*}|, \vnorm{\gtheta \str{x}{y}{\theta^*}},  \\
\vnorm{\gtheta^2 \str{x}{y}{\theta^*}}\Big\} \leq C.  
\end{aligned}
\end{equation*}
% \[
% \sup_{(x, y)\in\mathcal{X}\times \mathcal{Y}} \Big\{ |\ssf{x}{y}{\theta^*}|, \vnorm{\gtheta \ssf{x}{y}{\theta^*}}, \vnorm{\gtheta^2 \ssf{x}{y}{\theta^*}}\Big\} \leq C. 
% \]
where $\sigma_{\min}(\cdot)$ denotes the smallest singular value. 
  \label{assump:boundedness}
\end{assumption}
% We remark that when $|\mathcal{X}\times \mathcal{Y}|$ is finite, this assumption is automatically satisfied. 

% \subsubsection{Ranking}
\begin{theorem}[Ranking]
  Under Assumption~\ref{assump:ranking}, \ref{assump:identifiability-rank}, \ref{assump:compactness-rank}, \ref{assump:smoothness}, \ref{assump:boundedness}, there exists an integer $K_0$ such that for all $K\geq K_0$, as $n \rightarrow\infty$
  \begin{equation}
    \sqrt{n}\left(\wh{\theta}_R -\theta^* \right) ~\Rightarrow~\mathcal{N}(0, \mathcal{I}_{R, K}^{-1}), 
\end{equation}
for some matrix $\mathcal{I}_{R, K}$. There exists a constant C such that for all $K\geq K_0$, 
% and $s=\pm 1$, 
\begin{equation*}
\begin{aligned}
 |\mse_{\infty} (\wh{\theta}_R) -  \mse_{\infty} (\mle)| &\leq C/\sqrt{K}\\
\|\mathcal{I}_{R, K}^{-1} -  \fisher^{-1}\| &\leq C / \sqrt{K}
\end{aligned}
\end{equation*}
  \label{thm:ranking-mse}
\vspace{-0.4cm}
\end{theorem}

% {\em Proof.} See Section~\ref{sec:pf-ranking-mse}. 

\begin{theorem}[Binary]
Under Assumption~\ref{assump:binary}, \ref{assump:identifiability-binary}, \ref{assump:compactness-binary}, \ref{assump:smoothness}, \ref{assump:boundedness}, there exists an integer $K_0$ such that, for any $K\geq K_0$, as $n \rightarrow\infty$
  \begin{equation}
    \sqrt{n}\left(\wh{\theta}_B -\theta^* \right) ~\Rightarrow~\mathcal{N}(0, \mathcal{I}_{B, K}^{-1}), 
  \end{equation}
for some matrix $\mathcal{I}_{B, K}$. There exists a constant C such that for all $K\geq K_0$, 
% and $s=\pm 1$, 
\begin{equation*}
\begin{aligned}
|\mse_{\infty} (\wh{\theta}_B) -  \mse_{\infty} (\mle)| &\leq C /  K \\
\|\mathcal{I}_{B, K}^{-1} -  \fisher^{-1}\| &\leq C/K.  
\end{aligned}
\end{equation*}
  \label{thm:binary-mse}
  \vspace{-0.4cm}
\end{theorem}

\begin{remark}
  Theorem~\ref{thm:ranking-mse} and \ref{thm:binary-mse} reveal that under respective model assumptions, for any given $K\geq K_0$ both NCE estimators are asymptotically normal and $\sqrt{n}$-consistent. Moreover, both NCE estimators approach Fisher efficiency (statistical optimality) as $K$ grows. 
\end{remark}

\begin{table*}[t!]
\centering
\begin{small}
\begin{tabular}{|c|c|c|c|c|c|c|}
\hline
&   \multicolumn{2}{|c|}{Small}  &  \multicolumn{2}{|c|}{Medium}  &  \multicolumn{2}{|c|}{Large} \\
\hline
MLE  &  \multicolumn{2}{|c|}{111.5} & \multicolumn{2}{|c|}{82.7}   & \multicolumn{2}{|c|}{78.4}\\
\hline
 NCE   &Ranking &Binary         &Ranking &Binary   &Ranking &Binary   \\
 \hline
$K = 200$   & 113.8 & 106.8     & 83.2  & 82.1        & 79.3        &76.0   \\
\hline
$K = 400$   & 112.9 & 105.6     & 82.3  & 81.5        & 77.9      &75.6   \\
\hline
$K = 800$   & 111.9 & 105.3     & 81.4  & 81.6        & 77.8       &75.7   \\
\hline
$K = 1600$  & 110.6 & \textbf{104.8} & 81.7  & 81.5   & 77.5  &75.9   \\
\hline
reg-MLE      & \multicolumn{2}{|c|}{105.4}   &\multicolumn{2}{|c|}{79.9}     & \multicolumn{2}{|c|}{77.0} \\
\hline
reg-Ranking ($K = 1600$)  & \multicolumn{2}{|c|}{105.4}   &\multicolumn{2}{|c|}{\textbf{79.8}}     & \multicolumn{2}{|c|}{\textbf{75.0}} \\
\hline
reg-Binary ($K = 1600$)  & \multicolumn{2}{|c|}{\textbf{104.8}}  &\multicolumn{2}{|c|}{82.5}     & \multicolumn{2}{|c|}{75.7} \\
\hline
\end{tabular}
\end{small}
\caption{Perplexity on the test set of Penn Treebank. We show performance for
the ranking v.s. binary loss algorithms, with different values for $K$,
and with/without regularization.}
\label{table:lm-test}
\end{table*}

\begin{figure*}[t!]
\begin{minipage}{0.5\textwidth}
\includegraphics[width=\textwidth]{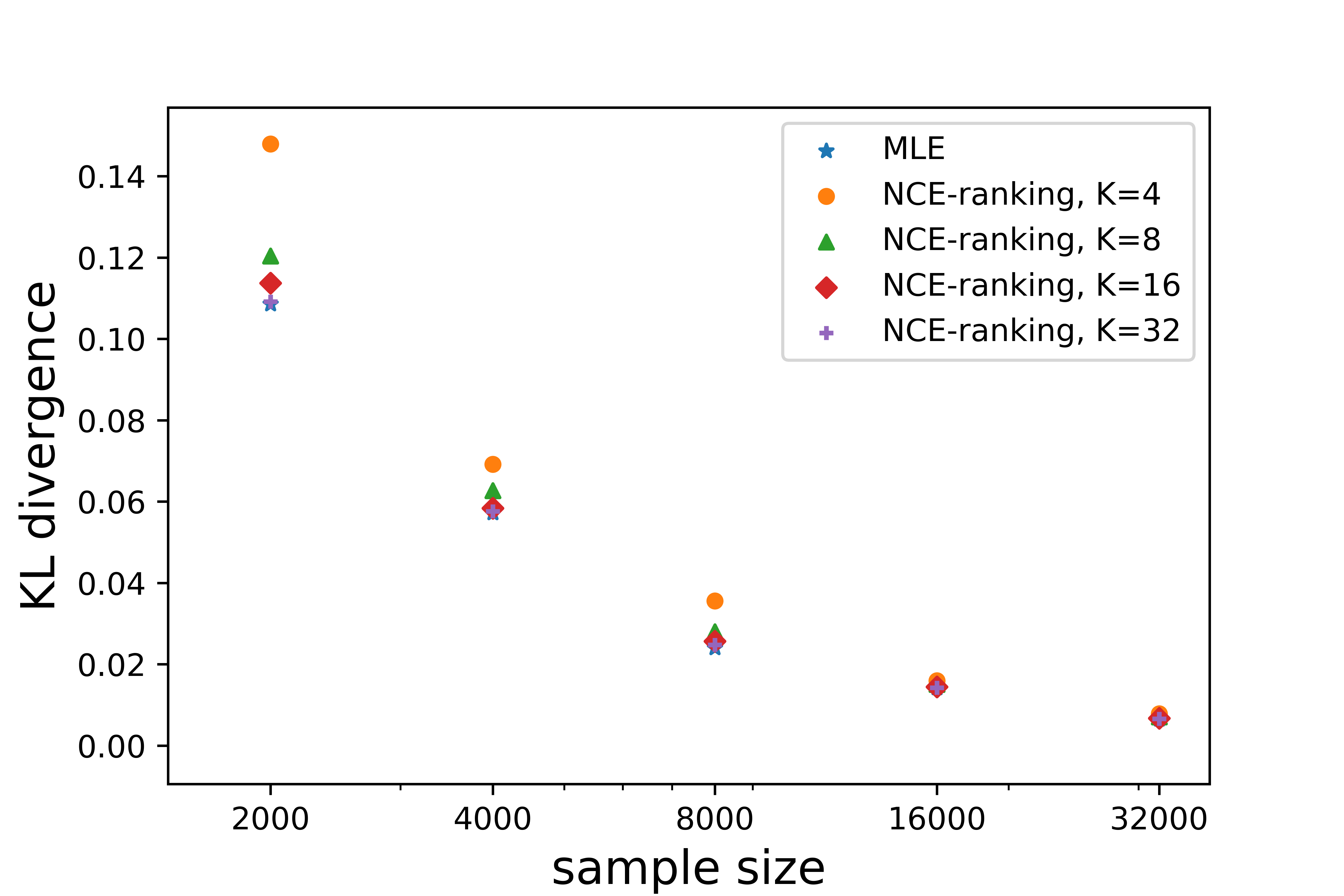}
\end{minipage}
\hspace{-1.5em}
% \hfill
\begin{minipage}{0.5\textwidth}
\includegraphics[width=\textwidth]{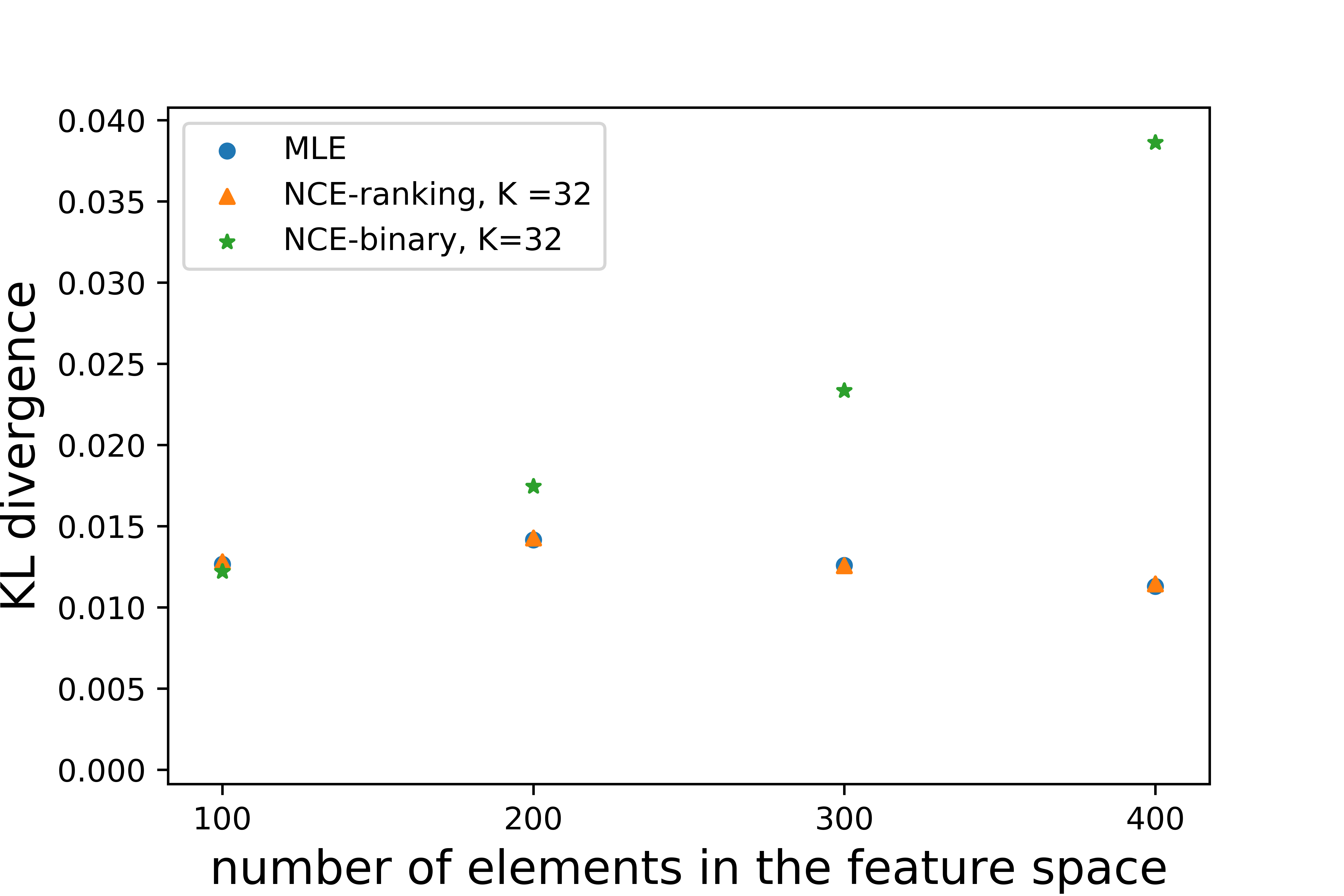}
\end{minipage}
\caption{KL divergence between the true distribution and the estimated distribution.}
\label{fig:sim}
\end{figure*}

\section{Experiments}
% In this section, we first present simulation study on a simple multi-class classification task where the true parameters are known to demonstrate the consistency and efficiency results implied by the theory. Then we evaluate the effectiveness of the two NCE algorithms on complex LSTM-based language models trained on Penn Treebank. 

\subsection{Simulations}
Suppose we have a feature space ${\cal X} \subset \mathbb{R}^d$ with $|{\cal X}| = m_x$, label space ${\cal Y} = \{1, \cdots, m_y \}$, and parameter $\theta = (\theta_1, \cdots, \theta_{m_y}) \in \mathbb{R}^{m_y\times d}$. Then for any given sample size $n$, we can generate observations ${(x^{(i)}, y^{(i)})}$ by first sampling $x^{(i)}$ uniformly from ${\cal X}$ and then sampling $y^{(i)}\in{\cal Y}$ by the condional model
\[
  p(y|x; \theta) = \exp(x^\prime \theta_y) / \sum_{y=1}^{m_y} \exp(x^\prime \theta_y).
\]
% \begin{equation}
%   p(y|x; \theta) = \exp(x^\prime \theta_y) / \sum_{y=1}^m \exp(x^\prime \theta_y), 
% \end{equation}
% where $\theta_y \in \mathbb{R}^d$ and $\theta = (\theta_1, \cdots, \theta_m) \in \mathbb{R}^{m\times d}$ is the parameter to be estimated. 
We first consider the estimation of $\theta$ by MLE and
NCE-ranking. We fix $d=4, m_x=200, m_y=100$ and generate ${\cal X}$
and the parameter $\theta$ from separate mixtures of Gaussians. We try different
configurations of $(n, K)$ and report the KL divergence between the
estimated distribution and true distribution, as summarized in the
left panel of figure~\ref{fig:sim}. The observations are:

$\bullet$ The NCE estimators are consistent for any fixed $K$. For a fixed sample size, the NCE estimators become comparable to MLE as $K$ increases. 

$\bullet$ The larger the sample size, the less sensitive are the NCE estimators to $K$. A very small value of $K$ seems to suffice for large sample size. 
% \begin{itemize}
%   \item The NCE estimators are consistent for any fixed $K$.  For a fixed sample size, the NCE estimators become comparable to MLE as $K$ increases.
%   \item The larger is the sample size, the less sensitive is NCE-ranking to $K$. For large sample size, very small value of $K$ suffices. 
%   % NCE-binary is more sensitive to $K$ than NCE-ranking. For large sample size, a small $K$ suffices for both NCE estimators to perform comparably to MLE.
% \end{itemize}

Apparently, under the parametrization above, the model is not self-normalized. To use NCE-binary, we add an extra $x$-dependent bias parameter $b_x$ to the score function (i.e. $s(x, y; \theta) = x^\prime\theta_y + b_x$) to make the model self-normalized or else the algorithm will not be consistent. Similar patterns to figure~\ref{fig:sim} are observed when varying sample size and $K$ (see Section~\ref{sec:sim-binary} of the supplementary material). However this makes NCE-binary not directly comparable to NCE-ranking/MLE since its performance will be compromised by estimating extra parameters and the number of extra parameters depends on the richness of the feature space ${\cal X}$. 
 To make this clear, we fix $n=16000, d=4, m_y=100, K=32$ and experiment with $m_x = 100, 200, 300, 400$. The results are summarized on the right panel of figure~2. As $|{\cal X}|$ increases, the KL divergence will grow while the performance of NCE-ranking/MLE is independent of $|{\cal X}|$. Without the $x$-dependent bias term for NCE-binary, the KL divergence will be much higher due to lack of consistency (0.19, 0.21, 0.24, 0.26 respectively). 

  % One thing worthing pointing out is, by adding a feature dependent bias term $b_x$, the estimation accuracy will depend on the richness of the feature space ${\cal X}$. 

% \begin{table*}[t!]
% \label{table:lm-test}
% \centering
% \begin{tabular}{|c|c|c|c|c|c|c|}
% \hline
%           &MLE &Ranking   &Binary &Binary     \\
%  \hline
% $n = 2000$   & 113.8 & 106.8     & 83.2  & 82.1        & 79.3        &76.0   \\
% \hline
% $n = 4000$   & 112.9 & 105.6     & 82.3  & 81.5        & 77.9      &75.6   \\
% \hline
% $n = 8000$   & 111.9 & 105.3     & 81.4  & 81.6        & 77.8       &75.7   \\
% \hline
% $n = 16000$  & 110.6 & \textbf{104.8} & 81.7  & 81.5   & 77.5  &75.9   \\
% \hline
% reg-MLE      & \multicolumn{2}{|c|}{105.4}   &\multicolumn{2}{|c|}{79.9}     & \multicolumn{2}{|c|}{77.0} \\
% \hline
% reg-Ranking ($K = 1600$)  & \multicolumn{2}{|c|}{105.4}   &\multicolumn{2}{|c|}{\textbf{79.8}}     & \multicolumn{2}{|c|}{\textbf{75.0}} \\
% \hline
% reg-Binary ($K = 1600$)  & \multicolumn{2}{|c|}{\textbf{104.8}}  &\multicolumn{2}{|c|}{82.5}     & \multicolumn{2}{|c|}{75.7} \\
% \hline
% \end{tabular}
% \caption{Perplexity on the test set. We show performance for
% the Ranking vs. Binary loss algorithms, with different values for $K$,
% and with/without regularization.}
% \end{table*}

\subsection{Language Modeling}
\label{sec:lm}

We evaluate the performance of the two NCE algorithms on a language
modeling problem, using the Penn Treebank (PTB) dataset
\cite{marcus1993building}. 
% The dataset has the standard
% split of 929k words for training, 73k words for validation, and 82k
% words for testing.  The vocabulary size is 10k. 
We choose \cite{zaremba2014recurrent} as the benchmark where the
 conditional distribution is modeled by two-layer LSTMs and the parameters are estimated by MLE (note that the current state-of-the-art is \cite{yang2018breaking}). 
\citet{zaremba2014recurrent} implemented 3 model configurations:
``Small'' , ``Medium'' and ``Large'', which have 200, 650 and 1500
units per layer respectively. We follow their setup
(model size, unrolled steps, dropout ratio, etc) but train the model by maximizing the two
NCE objectives. We use the unigram distribution as the negative sampling distribution and consider $K=200, 400, 800, 1600$. 
\commentout{
The current state-of-art results were achieved by \cite{zaremba2014recurrent}
using two-layer recurrent neural networks with LSTM units and dropout.
\cite{zaremba2014recurrent} implemented 3 model configurations:
``Small'' , ``Medium'' and ``Large'', which have 200, 650 and 1500
units per layer respectively (see \citep{zaremba2014recurrent} for
more details).  These models are trained by maximizing the likelihood
function, which is computationally intensive.  We follow the setup
(model size, unrolled steps, dropout ratio, etc) of
\cite{zaremba2014recurrent} but train the model by maximizing the two
NCE objectives. We use the unigram distribution as the negative sampling distribution and consider $K=200, 400, 800, 1600$. }
% and tune the learning rate, batch size and the maximum norm of the gradients.  

% The perplexities of the MLE on the Medium and Large models are directly copied from \citet{zaremba2014recurrent} but we obtained better results for MLE on the Small model (111.5), where \citet{zaremba2014recurrent} reported 114.5. 
The results on the test set are summarized in table~\ref{table:lm-test}. Similar patterns are observed on the validation set (see Section~\ref{sec:exp-validation} of the supplementary material). As shown in the table, the performance of NCE-ranking and NCE-binary improves as the number of negative examples increases, and finally outperforms the MLE. 

An interesting observation is, without regularization, the binary classification approach
outperforms both ranking and MLE. This suggests the model space (two-layer LSTMs) is rich enough as to approximately incorporate the $x$-dependent partition function $Z(\theta; x)$, thus making the model approximately self-normalized.
% in which case the binary approach covers the true model with a smaller model space than the ranking approach and MLE, and thus could achieve better finite-sample performance. 
\commentout{
Based on the theory in this paper, the success of the binary approach might indicate the
conditional distribution is approximately self-normalized.
When the intrinsic dimension of ${\cal X}$ is much smaller than its ambient dimension or the model complexity of the scoring function is large (e.g. neural networks), self-normalization could hold approximately, in which case the binary approach covers the true model with a smaller model space and hence could achieve better finite-sample performance. }
%, in which case both the ranking approach and the MLE search over a larger model space than the binary approach. 
This motivates us to modify the ranking
and MLE objectives by adding the following regularization term:
\begin{equation*}
\begin{aligned}
& \frac{\alpha}{n} \sum_{i=1}^n \left( \log \left( \frac{1}{m} \sum_{j=1}^{m} \exp \left( \ssf{x^{(i)}}{\wt{y}^{(i, j)}}{\theta} \right) \right) \right)^2  \\
& \approx \alpha \E_X \left[\left(\log Z(x; \theta) \right)^2\right], 
%\label{eq:regularize}
\end{aligned}
\end{equation*}
where $\wt{y}^{(i, j)}, 1\leq j\leq m$ are sampled from the noise distribution $p_N(\cdot)$. This regularization term promotes a constant partition
function, that is $Z(x; \theta)\approx 1$ for all $x\in\mathcal{X}$.
In our experiments, we fix $m$ to be 1/10 of the vocabulary size, $K =
1600$ and tune the regularization parameter $\alpha$.  As shown in the last three rows of the table,
regularization significantly improves the performance of both the ranking
approach and the MLE. 
% It does not improve the performance of the binary approach since the binary approach has implicitly assumed self-normalization. 
% Once regularization is included, the ranking approach outperforms the binary method for Medium and Large models.

\section{Conclusions}
In this paper we have analyzed binary and ranking variants of NCE for estimation of conditional models $p(y | x; \theta)$. The ranking-based variant is consistent for a broader class of models than the binary-based algorithm. Both algorithms achieve Fisher efficiency as the number of negative examples increases. 
Experiments show that both algorithms outperform MLE on a language modeling task.  
The ranking-based variant of NCE outperforms the binary-based variant once a regularizer is introduced that encourages self-normalization.

\section*{Acknowledgments}

The authors thank Emily Pitler and Ali Elkahky for many useful conversations about the work, 
and David Weiss for comments on an earlier draft of the paper.

% \section*{Acknowledgments}

% The acknowledgments should go immediately before the references.  Do
% not number the acknowledgments section. Do not include this section
% when submitting your paper for review. \\

\bibliography{nce}
\bibliographystyle{acl_natbib_nourl}

 %!TEX root = emnlp2018_all.tex
\appendix
\section{Experiment Results}
\subsection{Similation Results for NCE-binary}
\label{sec:sim-binary}
\begin{figure}[h]
\includegraphics[width=0.5\textwidth]{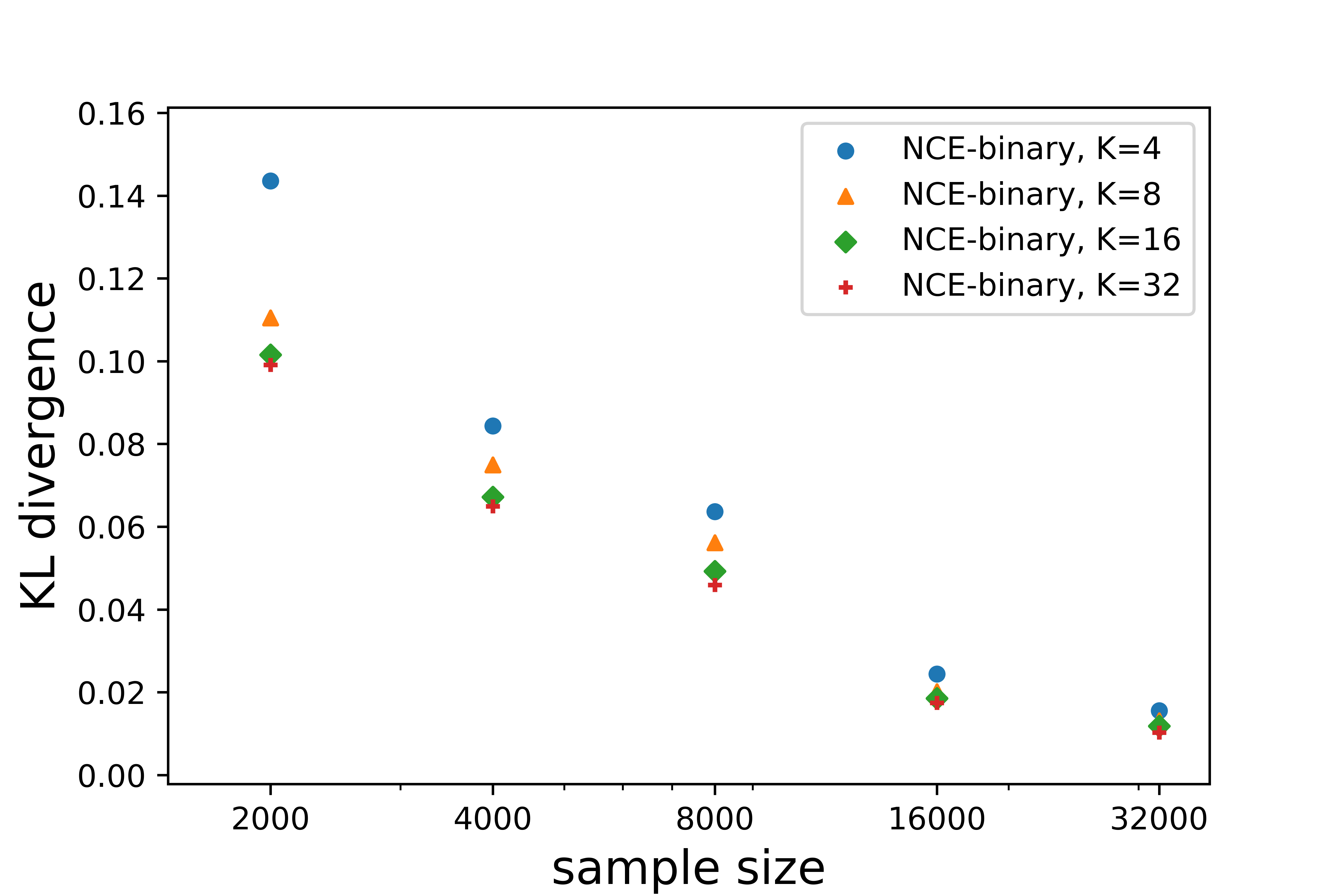}
\caption{KL divergence between the true distribution and the estimated distribution.}
\label{fig:sim-binary}
\end{figure}

See figure~\ref{fig:sim-binary}.

\subsection{Perplexity on Validation Set of Penn Treebank}
\label{sec:exp-validation}
See table~\ref{table:lm-validation} below.
\begin{table*}[t!]
\centering
\begin{tabular}{|c|c|c|c|c|c|c|}
\hline
&   \multicolumn{2}{|c|}{Small}  &  \multicolumn{2}{|c|}{Medium}  &  \multicolumn{2}{|c|}{Large} \\
\hline
MLE
  &  \multicolumn{2}{|c|}{116.8} & \multicolumn{2}{|c|}{86.2}   & \multicolumn{2}{|c|}{82.2}\\
  \hline
 NCE   &Ranking &Binary     &Ranking &Binary   &Ranking &Binary   \\
 \hline
$K = 200$   & 121.1 & 112.1     & 87.8  & 86.3    & 84.1  &80.8   \\
\hline
$K = 400$   & 119.0 & 110.6     & 87.2  & 85.8    & 82.9  &80.0   \\
\hline
$K = 800$   & 116.8 & 110.1     & 86.4  & 85.5    & 82.3  &79.9   \\
\hline
$K = 1600$  & 116.5 & \textbf{109.8} & 86.3  & 85.4   & 82.0  &79.5   \\
\hline
reg-MLE      & \multicolumn{2}{|c|}{109.9}   &\multicolumn{2}{|c|}{\textbf{83.1}}     & \multicolumn{2}{|c|}{79.9} \\
\hline
reg-Ranking ($K = 1600$)  & \multicolumn{2}{|c|}{110.5}   &\multicolumn{2}{|c|}{83.3}     & \multicolumn{2}{|c|}{\textbf{78.4}} \\
\hline
reg-Binary ($K = 1600$)  & \multicolumn{2}{|c|}{109.0}  &\multicolumn{2}{|c|}{86.1}     & \multicolumn{2}{|c|}{79.7} \\
\hline
\end{tabular}
\caption{Perplexity on the validation set. We show performance for
the Ranking vs. Binary loss algorithms, with different values for $K$,
and with/without regularization.}
\label{table:lm-validation}
\end{table*}

\section{Proofs}
\label{sec:supplemental}
Recall we have the following setup: 
\begin{itemize}
  \item $\mathcal{X}, \mathcal{Y}$ are finite sets
  \item $p_X(x), p_Y(y), p_N(y) > 0, \forall x\in\mathcal{X}, y\in\mathcal{Y}$.
  % \item $p_N(y) > 0$ for any $y$ such that $p_Y(y) > 0$
\end{itemize}

We present the proofs for ranking loss and binary classification loss separately, in the following two subsections. 

The following fact will be frequently used in the consistency proof. 
\begin{lemma}
Define the $K$-simplex by 
\[
\mathcal{D}_K = \left\{x | x\in\reals_+^{K+1},~ \sum_{i=0}^K x_i =1 \right\}. 
\]
Then for any $p\in\mathcal{D}_K$, 
\[
p = \arg\max_{q\in\mathcal{D}_K} \sum_{i=0}^K q_i\log p_i.
\]
\label{lem:entropy}
\end{lemma}
% These two assumptions immediately imply that: 

\subsection{Proofs for Ranking Loss}
\label{sec:proof-ranking}
To simplify notations, we use $y_{0:K}$ to represent the vector $(y_0, \cdots, y_K)$ and $y^{(i, 0:K)}$ to represent $(y^{(i, 0)}, \cdots, y^{(i, K)})$ where we define $y^{(i, 0)}:= y^{(i)}$.

For any tuple $(x, y_{0:K})$ and parameter $\theta$, we define 
\[
q(i | x, y_{0:K}; \theta)
= \frac{\exp( \ssf{x}{y_i}{\theta})}{\sum_{i=0}^K \exp( \ssf{x}{y_i}{\theta})}.
\]

% \begin{lemma}
% If the scoring function $\ssf{x}{y}{\theta}$ is twice continuous differentiable w.r.t. $\theta$ for all $(x, y)\in\mathcal{X}\times \mathcal{Y}$ where $|\mathcal{X}|, |\mathcal{Y}|<\infty$, then for any given compact sets $\Theta\subset\reals^d, \Gamma\subset\reals$, there exists constant $C$ such that, 
% \begin{equation*}
% \begin{aligned}
% \sup_{x, y, \theta^*\in\Theta, \gamma^*\in\Gamma} \Big\{ \exp \left(|\ssf{x}{y}{\theta^*}| + |\gamma^*|\right), \vnorm{\gtheta \ssf{x}{y}{\theta^*}}, \vnorm{\gtheta^2 \ssf{x}{y}{\theta^*}}\Big\} \leq C
% \end{aligned}
% \end{equation*}
% \label{lem:uniform-bound}
% \end{lemma}

\begin{lemma}
% Define
% \[
% q(i | x, y_{0:K}; \theta)
% = \exp( \ssf{x}{y_i}{\theta})/\sum_{i=0}^K \exp( \ssf{x}{y_i}{\theta})
% \]
% Then 
Under Assumption~\ref{assump:ranking}, 
\[
q(i | x, y_{0:K}; \theta^*)
= \frac{p_{Y|X}(y_i|x)/p_N(y_i)}{\sum_{j=0}^K p_{Y|X}(y_j|x)/p_N(y_j)}
\]
\label{lemma:qi}
\end{lemma}

{\em Proof:} The lemma follows Assumption~\ref{assump:ranking} that
\begin{equation*}
\begin{aligned}
p_{Y|X}(y | x) & = \frac{ \exp(\str{x}{y}{\theta^*})}{Z(x; \theta^*)} \\
& = \frac{ p_N(y)\exp(\ssf{x}{y}{\theta^*})}{Z(x; \theta^*)}
\end{aligned}
\end{equation*}
$\qed$

\begin{lemma}
Define 
% $q(i | x, y_{0:K}; \theta)$ as in Lemma~\ref{lemma:qi}, and
\[
p(y | x; \theta) = \frac{ \exp \{ \str{x}{y}{\theta} \}} { \sum_y
   \exp \{ \str{x}{y}{\theta} \} } 
\]
For any $\theta_1, \theta_2$, the following three statements are equivalent: 
\begin{enumerate}[label={(\arabic*)}]
  \item  For all $x, y$ 
  \[
  p(y | x; \theta_1) = p(y | x; \theta_2)
  \]
  \item  For all $x, y_{0:k}$ and $0\leq i\leq K$ 
  \[
  q(i | x, y_{0:k}; \theta_1) = q(i | x, y_{0:k}; \theta_2)
  \]
  \item  There exists some function $c(\cdot)$ such  that for all $x, y$
  \[
  \str{x}{y}{\theta_1} - \str{x}{y}{\theta_2} = c(x).
  \]
\end{enumerate}
\label{lem:iden}
\end{lemma}

{\em Proof:} 
By Lemma~\ref{lemma:qi},
\[
  q(i | x, y_{0:k}; \theta)  = \frac{p(y_i | x; \theta) / p_N(y_i)}{\sum_{i=0}^K p(y_i | x; \theta) / p_N(y_i)  },
\]
and hence $(1)\Rightarrow(2)$. 
To show $(2)\Rightarrow (3)$, notice that, for any $x, y, y'$, by setting $i = 0$ and $y_0 = y$, $y_1 = y_2 \ldots = y_K = y'$, $q(i | x, y_{0:k}; \theta_1) = q(i | x,
y_{0:k}; \theta_2)$ will imply
% for all $x, y_{0:k}, i$,
\begin{equation*}
\begin{aligned}
&  \frac{\exp( \ssf{x}{y}{\theta_1} ) }{\exp ( \ssf{x}{y}{\theta_1} ) + K \exp ( \ssf{x}{
y'}{\theta_1} )} \\
=  & \frac{\exp( \ssf{x}{y}{\theta_2} )}{\exp( \ssf{x}{y}{\theta_2} ) + K \exp( \ssf{x}{y'}{\theta_2} )},
\end{aligned}
\end{equation*}
or equivalently,
\begin{equation*}
\begin{aligned}
& \frac{ 1 }{ 1 + K \exp ( \ssf{x}{
y'}{\theta_1} -  \ssf{x}{y}{\theta_1} )} \\
= &  \frac{ 1 }{ 1 + K \exp ( \ssf{x}{
y'}{\theta_2} -  \ssf{x}{y}{\theta_2} )}. 
\end{aligned}
\end{equation*}
Then it follows that, 
\[
   \ssf{x}{y'}{\theta_1} - \ssf{x}{y}{\theta_1} = 
 \ssf{x}{y'}{\theta_2} - \ssf{x}{y'}{\theta_2},
\]
and therefore, 
\[
  \str{x}{y}{\theta_1} - \str{x}{y}{\theta_2} =
\str{x}{y'}{\theta_1} - \str{x}{y'}{\theta_2}.
\]
So $\str{x}{y}{\theta_1} - \str{x}{y}{\theta_2}$ only depends on $x$. 
Finally, we show $(3)\Rightarrow (1)$. If (3) holds, for all $x, y, y'$
\[
   \str{x}{y}{\theta_1} - \str{x}{y}{\theta_2} = \str{x}{y'}{\theta_1} - \str{x}{y'}{\theta_2},
\]
which implies
\[
  \str{x}{y}{\theta_1} - \str{x}{y'}{\theta_1} =
\str{x}{y}{\theta_2} - \str{x}{y'}{\theta_2}.
\]
We then have 
% \begin{eqnarray*}
% p(y | x; \theta_1) 
% &=& \frac{ \exp(\str{x}{y}{\theta_1})}{\sum_{y'}
%   \exp(\str{x}{y'}{\theta_1})}\\
% &=& \frac{1}{\sum_{y'}
%   \exp(\str{x}{y'}{\theta_1} - \str{x}{y}{\theta_1})}\\
% &=& \frac{1}{\sum_{y'}
%   \exp(\str{x}{y'}{\theta_2} - \str{x}{y}{\theta_2})}\\
% &=& p(y | x; \theta_2). 
% \end{eqnarray*}
\begin{equation*}
  \begin{aligned}
    p(y | x; \theta_1)  &=  \frac{ \exp(\str{x}{y}{\theta_1})}{\sum_{y'}
  \exp(\str{x}{y'}{\theta_1})}\\
&= \frac{1}{\sum_{y'}
  \exp(\str{x}{y'}{\theta_1} - \str{x}{y}{\theta_1})}\\
&= \frac{1}{\sum_{y'}
  \exp(\str{x}{y'}{\theta_2} - \str{x}{y}{\theta_2})}\\
&= p(y | x; \theta_2). 
  \end{aligned}
\end{equation*}
$\qed$

\subsubsection{Proof of Theorem~\ref{thm:pop-consistency-rank}}
\label{sec:pf-pop-consis-ranking}
If we define 
\begin{eqnarray*}
&&g_i(\theta; x, y_{0:K})\\ 
&=& p_{X, Y}(x, y_i) \prod_{j \neq i} p_N(y_j) \log
q(i | x, y_{0:K}; \theta) 
\end{eqnarray*}
then for any $i \in \{0 \ldots K\}$,
we have
\[
L_R^\infty\left(\theta\right) = \sum_{x}\sum_{y_{0:K}} g_i(x, y_{0:K}; \theta)
\]
It follows that 
\begin{eqnarray*}
L_R^\infty\left(\theta\right) 
&=& \frac{1}{K+1} \sum_{i=0}^K \sum_{x}\sum_{y_{0:K}} g_i(x, y_{0:K}; \theta)\\
&=& \frac{1}{K+1} \sum_{x}\sum_{y_{0:K}} \sum_{i=0}^K g_i(x, y_{0:K}; \theta)\\
&=& \frac{1}{K+1}\sum_{x}\sum_{y_0,
  y_1,\cdots, y_K}f(\theta ; x, y_{0:K}) 
\end{eqnarray*}
where
\begin{equation*}
\begin{aligned}
& f(\theta; x, y_{0:K}) = \sum_{i=0}^K g_i(x, y_{0:K}; \theta)\\
& = \sum_{i=0}^K p_{X, Y}(x, y_i)\prod_{j\neq i} p_N(y_j) \log q(i | x, y_{0:K}; \theta)\\
& = \left(\sum_{i=0}^K p_{X, Y}(x, y_i)\prod_{j\neq i} p_N(y_j)\right) \times \sum_{i=0}^K \\
&\frac{p_{X, Y}(x, y_i)\prod_{j\neq i} p_N(y_j)}{\sum_{i=0}^K p_{X, Y}(x, y_i)\prod_{j\neq i} p_N(y_j)} \log q(i | x, y_{0:K}; \theta)\\
& = \left(\sum_{i=0}^K p_{X, Y}(x, y_i)\prod_{j\neq i} p_N(y_j)\right) \times \\
& \sum_{i=0}^K \frac{p_{Y|X}(y_i|x) / p_N(y_i)}{\sum_{j=0}^K p_{Y|X}(y_j|x) / p_N(y_j)} \log q(i | x, y_{0:K}; \theta).  
\end{aligned}
\end{equation*}
Next note that $\sum_{i=0}^K q(i | x, y_{0:K}; \theta) = 1$, hence by
Lemmas~\ref{lem:entropy} and~\ref{lemma:qi}, $\theta^* \in
\arg\max_{\theta} f(\theta; x, y_{0:K})$ for all $x, y_{0:K}$, hence
$\theta^* \in \arg\max_{\theta} L^{\infty}_R(\theta)$.

It follows that for any $\bar{\theta} \in \arg\max_{\bar{\theta}}
L^{\infty}_R(\bar{\theta})$, $\bar{\theta}$ must be in $\arg\max_{\theta}
f(\theta; x, y_{0:K})$ for all $x, y_{0:K}$, from which it follows
that 
\[
q(i | x, y_{0:k}; \bar{\theta}) = q(i | x, y_{0:k}; \theta^*)
\]
for all $x, y_{0:k}, i$. Then by Lemma~\ref{lem:iden},
for all $x, y$
\[
p(y | x; \bar{\theta}) = p(y | x; \theta^*) = p_{Y|X}(y | x).
\]
This proves the first part of the theorem. To prove the second part, for any $\bar{\theta}\in\Theta^*_R$, by the first part of this theorem, for all $x, y$
\[
p(y | x; \bar{\theta}) = p(y | x; \theta^*) = p_{Y|X}(y | x).
\]
It follows from Lemma~\ref{lem:iden} that there exists function $c(x)$ such that for all $x, y$
\[
\str{x}{y}{\bar{\theta}} - \str{x}{y}{\theta^*} = c(x).
\]
If Assumption~\ref{assump:identifiability-rank} holds, $\theta = \theta^*$, that is $\Theta_R^*$ is a singleton. 
On the other hand, if there exists function $c(x)$ and parameter $\bar{\theta}$ such that for all $x, y$
\[
\str{x}{y}{\bar{\theta}} - \str{x}{y}{\theta^*} = c(x), 
\]
then by Lemma~\ref{lem:iden}, for all $x, y$
\[
p(y | x; \bar{\theta}) = p(y | x; \theta^*) = p_{Y|X}(y | x).
\]
So $\theta \in \Theta^*_R$. If $\Theta^*_R$ is a singleton, it must be the case that $\bar{\theta} = \theta^*$. This proves the second part of the theorem. 

$\qed$

\subsubsection{Proof of Theorem~\ref{thm:consistency-rank}}
\label{sec:pf-consis-ranking}
Under the assumptions in Theorem~\ref{thm:consistency-rank}, by classical large sample theory (e.g. Theorem 16(a) in \cite{ferguson1996course}), for any compact set $\mathcal{K}\subset \Theta$,
\begin{equation}
  \mathbb{P} \left\{\lim\limits_{n\rightarrow\infty}\sup\limits_{\theta\in\mathcal{K}} |\objr(\theta) - L_R^\infty(\theta)| = 0 \right\} = 1
  \label{eq:uniform-convergence}
\end{equation}
Since $|\objr(\theta) - L_R^\infty(\theta)|\geq \objr(\theta) - L_R^\infty(\theta)$, 
\begin{equation}
    \mathbb{P} \left\{\limsup\limits_{n\rightarrow\infty}\sup\limits_{\theta\in\mathcal{K}} \left(\objr(\theta) - L_R^\infty(\theta)\right) \leq 0 \right\} = 1. 
    \label{eq:consistency-1}
\end{equation}
For any $\wh{\theta}_{\mathcal{K}}^n \in \arg\max\limits_{\theta\in\mathcal{K}}\objr(\theta)$, 
\begin{align*}
\sup\limits_{\theta\in\mathcal{K}} & \left(\objr(\theta) - L_R^\infty(\theta)\right) \geq \objr(\wh{\theta}_{\mathcal{K}}^n) - L_R^\infty(\wh{\theta}_{\mathcal{K}}^n) \\
& \geq \objr(\wh{\theta}_{\mathcal{K}}^n) - \sup\limits_{\theta\in\mathcal{K}} L_R^\infty(\theta) \\
& = \sup\limits_{\theta\in\mathcal{K}} \objr(\theta) - \sup\limits_{\theta\in\mathcal{K}} L_R^\infty(\theta).
\end{align*}
Combining this inequality and \eqref{eq:consistency-1} yields
\begin{equation}
    \mathbb{P} \left\{\limsup\limits_{n\rightarrow\infty}\sup\limits_{\theta\in\mathcal{K}} \objr(\theta) \leq \sup\limits_{\theta\in\mathcal{K}}L_R^\infty(\theta)  \right\} = 1.
    \label{eq:strong1}
\end{equation}
Recall that 
\[
\Theta^*_R = \left\{\bar{\theta}~|~\bar{\theta} \in\arg\max_{\theta\in\Theta} L_R^\infty(\theta)\right\}.
\]
For any $\rho > 0$, define 
\[
\Theta_{\rho} = \left\{\theta~|~\theta\in\Theta, ~\arg\min_{\theta^*\in\Theta^*_R} \|\theta - \theta^*\| \geq \rho\right\}.
\]
By Assumption~\ref{assump:continuity}, $L_R^\infty(\theta)$ is a continuous function of $\theta$ and therefore, 
\[
\delta_{\rho} \doteq \sup_{\theta\in\Theta_{\rho}}L_R^\infty(\theta)<\delta\doteq \sup_{\theta\in\Theta}L_R^\infty(\theta). 
\]
Apply the uniform convergence results in equation~\eqref{eq:strong1} with $\mathcal{K} = \Theta_{\rho}$, with probability 1, 
\begin{equation}
  \limsup\limits_{n\rightarrow\infty}\sup\limits_{\theta\in\Theta_{\rho}} \objr(\theta) \leq \delta_{\rho} < \delta. 
  \label{eq:consis-ranking-1}
\end{equation}
On the other hand, substitute $|\objr (\theta) - L_R^\infty(\theta)|\geq L_R^\infty(\theta) - \objr(\theta)$ into~\eqref{eq:uniform-convergence}, with the same arguments as used to obtain~\eqref{eq:strong1}, one could also show that, for any compact set $\mathcal{K}\subset \Theta$,
\[
    \mathbb{P} \left\{ 
    \liminf\limits_{n\rightarrow\infty} \sup\limits_{\theta\in\mathcal{K}} \objr (\theta)  \geq \sup \limits_{\theta\in\mathcal{K}} L_R^\infty(\theta)
    \right\} = 1.
\]
Let $\mathcal{K} = \Theta$, by definition, 
\begin{equation}
  \mathbb{P} \left\{ \liminf\limits_{n\rightarrow\infty} \objr(\wh{\theta}_R) \geq  \delta \right\} = 1.
    \label{eq:consis-ranking-2}
\end{equation}
Combine equations~\eqref{eq:consis-ranking-1}, \eqref{eq:consis-ranking-2}, there exists integer $N$ such that 
\[
    \mathbb{P} \left\{ \wh{\theta}_R^n \not\in \Theta_{\rho}, ~\text{for all}~ n\geq N\right\} = 1.
\]
Since this holds for any $\rho > 0$,  
\begin{equation}
  \mathbb{P} \Big\{\lim_{n\rightarrow \infty} \min_{\theta^*\in\Theta^*_R}\| \wh{\theta}_R^n - \theta^*\| = 0 \Big\}.
  \label{eq:est-consistency-1}
\end{equation}
Define the mapping $g$ from the parameter space $\Theta$ to the function space on $\mathcal{X}\times\mathcal{Y}$ by $g(\theta) = p(\cdot, \cdot~; \theta)$ where
\[
p(x, y; \theta) = \exp(\str{x}{y}{\theta})/Z(x; \theta),
\]
and $Z(x; \theta) = \sum_{y\in\mathcal{Y}}\exp(\str{x}{y}{\theta})$. 
By Theorem~\ref{thm:pop-consistency-rank}, 
\begin{equation}
  \Theta^*_R = \left\{\theta~|~ g(\theta) = p_{Y|X}, \theta\in\Theta\right\}.
  \label{eq:est-consistency-2}
\end{equation}
Further by Assumption~\ref{assump:continuity}, $g(\theta)$ is a continuous function of $\theta$ under the metric $d(\cdot, \cdot)$. Since $\Theta$ is compact set, $g(\theta)$ is uniform continuous. By uniform continuity, for any $\varepsilon > 0$, there exists $\Delta$ such that, for any $\theta_1, \theta_2$ satisfying $\|\theta_1 - \theta_2\|\leq \Delta$, 
\[
d \left(g(\theta_1), g(\theta_2)\right) \leq \varepsilon. 
\]
Hence, equation~\eqref{eq:est-consistency-1} implies
\[
    \mathbb{P} \left\{ \lim\limits_{n\rightarrow \infty} d \left(\wh{p}_{Y|X}^n, p_{Y|X}\right) = 0\right\} = 1. 
\]
When assumption~\ref{assump:identifiability-rank} holds, by Theorem~\ref{thm:pop-consistency-rank},  $\Theta^* = \{\theta^*\}$ is a singleton and equation~\eqref{eq:est-consistency-1} is reduced to
\[
    \mathbb{P} \left\{\lim\limits_{n\rightarrow \infty} \wh{\theta}_R^n = \theta^*\right\} = 1. 
\]

\subsubsection{Proof of Theorem~\ref{thm:ranking-mse}}
\label{sec:pf-ranking-mse}
Since we have defined $y^{(i, 0)} := y^{(i)}$, the objective function $\objr(\theta )$ can be written as the average of the following I.I.D. random variables $ L^{(i)}_R(\theta), 1\leq i \leq n$:  
\begin{equation}
\begin{aligned}
\objr(\theta ) &= \frac{1}{n}\sum_{i=1}^n  \log\left(\frac{\exp( \ssf{x^{(i)}}{y^{(i, 0)}}{\theta} )}{\sum_{j=0}^K \exp( \ssf{x^{(i)}} {y^{(i, j)}} {\theta} }\right) \\
&  := \frac{1}{n}\sum_{i=1}^n L^{(i)}_R(\theta). 
\end{aligned}
\end{equation}
Define 
\[
L_R(\theta) = \log \left( \frac{\exp(\ssf{x}{y_0}{\theta})}{\sum_{j=0}^K \exp(\ssf{x}{y_j}{\theta}) ) }\right)
\]
where $(x, y_{0:K}) \sim p_{X, Y}(x, y_0)\prod_{j=1}^K p_N(y_j)$. Then we have 
\[
(x^{(i)}, y^{(i, 0:K)}) =_d (x, y_0, \cdots, y_K) 
\]
\[
L^{(i)}_R(\theta) =_d L_R(\theta)
\]
for $1\leq i \leq n$, where $=_d$ represents equal in distribution. 

\begin{lemma}
For any $g(x, y_0, \cdots, y_K): \mathcal{X}\times\mathcal{Y}^{K+1} \rightarrow \mathbb{R}$ satisfying  
$$g(x, y_0, \cdots, y_K) = g(x, y_{\pi(0)}, \cdots, y_{\pi(K)}),$$
where $\pi$ is any permutation on $\{0, 1, \cdots, K\}$, the following equality holds for any function 
$f(\cdot, \cdot): \mathcal{X}\times\mathcal{Y} \rightarrow \mathbb{R}^l$ where $l\in\mathbb{N}_+$.
\begin{equation}
\begin{aligned}
  & \E \left[  \sum_{j=0}^K \q{j} g(x, y_0, \cdots, y_K)  \right] 
  \\
  = & \E \left[g(x, y_0, \cdots, y_K)f(x, y_0) \right], 
\end{aligned}
\end{equation}
where $(x, y_{0:K}) \sim p_{X, Y}(x, y_0)\prod_{j=1}^K p_N(y_j)$. 
% \[
% (x, y_0, \cdots, y_K)\sim p_{X, Y}(x, y_0)\prod_{j=1}^K p_N(y_j).
% \]
\label{lem:integral-rank}
\end{lemma}

{\em Proof: }
We use $g(x, y_{0:K})$ to represent $g(x, y_0, \cdots, y_K)$. By definition, 
\begin{equation*}
\begin{aligned}
    & \E \left[  \sum_{j=0}^K \q{j} g(x, y_{0:K})  \right]\\
   = & \sum_{x,y_{0:K}} p_{X,Y}(x, y_0; \theta) \prod_{j=1}^K p_N(y_j) \times \\
   &  \sum_{j=0}^K \q{j} g(x, y_{0:K}) \\
   = & \sum_{x,y_{0:K}} p_{X}(x) \prod_{j=0}^K p_N(y_j) \frac{p_{Y|X}(y_0|x; \theta)}{p_N(y_0)}\times \\
   & \sum_{j=0}^K \q{j} g(x, y_{0:K})
\end{aligned}
\end{equation*}
By symmetry, 
\begin{equation*}
\begin{aligned}
   & \E \left[  \sum_{j=0}^K \q{j} g(x, y_{0:K})  \right]\\
   = & \sum_{x,y_{0:K}}  \frac{p_{X}(x)\prod_{j=0}^K p_N(y_j) }{K+1} \sum_{l=0}^{K}\frac{p_{Y|X}(y_l|x; \theta)}{p_N(y_l)} \\
   & \times  \sum_{j=0}^K \q{j} g(x, y_{0:K})
\end{aligned}
\end{equation*}
Plug in the definition of $\q{j}$ and by symmetry, 
\begin{equation*}
\begin{aligned}
   & \E \left[  \sum_{j=0}^K \q{j} g(x, y_{0:K})  \right]\\
   = & \sum_{x,y_{0:K}} \Big\{ \frac{p_{X}(x)\prod_{j=0}^K p_N(y_j) }{K+1} \times \\
   &  \sum_{j=0}^K \frac{p_{Y|X}(y_j|x)}{p_N(y_j)} g(x, y_{0:K})\Big\} \\
   = & \sum_{x,y_{0:K}} p_{X}(x) \prod_{j=0}^K p_N(y_j) \frac{p_{Y|X}(y_0|x)}{p_N(y_0)} \\ 
   & \times g(x, y_{0:K})f(x, y_0)\\
  = & \E \left[g(x, y_{0:K})f(x, y_0) \right]
\end{aligned}
\end{equation*}
$\qed$

\begin{lemma}
  Under the assumptions in Theorem~\ref{thm:ranking-mse}, as $n\rightarrow\infty$
\begin{equation*}
\begin{aligned}
 \sqrt{n}\left(\wh{\theta}_R - \theta^*\right) \rightarrow  \mathcal{N}\left(0, V_R\right) 
\end{aligned}
\end{equation*}
where  $V_R = $
\[
\E \left[\gtheta^2 L_R(\theta^*)\right]^{-1} \var \left[\gtheta L_R(\theta^*) \right] \E \left[\gtheta^2 L_R(\theta^*)\right]^{-1}
\]
\label{lem:asym-rank}
\end{lemma}
{\em Proof:}
By the Mean-Value Theorem,
\begin{equation*}
\begin{aligned}
& \gtheta \objr(\wh{\theta}_R) = \gtheta \objr(\theta^*)  + \\
& \left[\int_0^1 \gtheta^2 \objr\left(\theta^* + t(\wh{\theta}_R-\theta^*)  \right)dt\right] (\wh{\theta}_R - \theta^*). 
\end{aligned}
\end{equation*}
Notice that $\wh{\theta}_R$ is the maximizer of $\objr(\theta)$. By Theorem~\ref{thm:consistency-rank}, $\wh{\theta}_R$ is strongly consistent. Since $\theta^*$ is in the interior of a compact set $\Theta$, there exists integer $N$ such that for all $n > N$ with probability 1, $\wh{\theta}_R$ is also in the interior of $\Theta$. So $\gtheta \objr(\wh{\theta}_R) = 0 $ and therefore,
\begin{equation*}
\begin{aligned}
\wh{\theta}_R - \theta^* & =  -\left[\int_0^1 \gtheta^2 \objr\left(\theta^* + t(\wh{\theta}_R-\theta^*)  \right)dt\right]^{-1} \\ 
& \times \gtheta \objr(\theta^*) 
\end{aligned}
\end{equation*}
On the one hand, by Strong Law of Large Numbers, for any $\theta$, almost surely,
\[
\gtheta^2 \objr (\theta) \rightarrow  \E [ \gtheta^2 L_R(\theta) ].
\]
On the other hand, by Theorem~\ref{thm:consistency-rank}, $\wh{\theta}_R$ is strongly consistent. So for any $t\in [0, 1]$
 \begin{equation*}
    \gtheta^2 \objr\left(\theta^* + t(\wh{\theta}_R-\theta^*)  \right) \rightarrow \E[\gtheta^2 L_R(\theta^*)].
  %\label{eq:hessian-converge-rank}
\end{equation*}
Also notice that, 
\begin{equation}
\begin{aligned}
 \gtheta L_R & (\theta^*) = \gtheta  \ssf{x}{y_0}{\theta^*} -  \\
 & \sum_{j=0}^K q( j | x, y_{0:K}; \theta^*) \gtheta \ssf{x}{y_j}{\theta^*}.
\label{eq:first-derivative-rank}
\end{aligned}
\end{equation}
% where $q( j | x^{(i)}, y^{(i, 0:K)}; \theta^*) = \exp(s(x^{(i)}, y^{(i, j)}; \theta^*))/ \sum_{l=0}^K \exp(s (x^{(i)}, y^{(i, l)}; \theta^*))$. 
By Lemma~\ref{lem:integral-rank}, 
\[
\E [L_R(\theta^*)] = 0.
\]
Hence $\E [\objr (\theta^*)] = 0$ and by Central Limit Theorem, 
\begin{equation}
    \sqrt{n}\gtheta \objr(\theta^*)\rightarrow \mathcal{N} \left(0, \var \left[\gtheta L_R(\theta^*) \right]\right). 
  \label{eq:asymp-normal-classify}
\end{equation}
Combining these facts together, 
\[
  \sqrt{n}\left(\wh{\theta}_R - \theta^*\right) \rightarrow \mathcal{N}\left(0, V_R \right) 
\]
$\qed$

Define the matrix $W_{R, K}$ by 
\begin{equation*}
\begin{aligned}
& W_{R, K} \\
= & \E \left[ \left(\sum_{j=0}^K q( j | x, y_{0:K}; \theta^*) \gtheta \ssf{x}{y_j}{\theta^*}\right) \times \right. \\
& \left. \left(\sum_{j=0}^K q( j | x, y_{0:K}; \theta^*) \gtheta \ssf{x}{y_j}{\theta^*}\right)^\top\right], 
\end{aligned}
\end{equation*}
where the expectation is w.r.t. $(x, y_0, \cdots, y_K)\sim p_{X, Y}(x, y_0)\prod_{j=1}^K p_N(y_j)$. 

\begin{lemma}
% With $(x, y_{0:K})\sim p_{X, Y}(x, y_0)\prod_{j=1}^K p_N(y_j)$, 
\begin{equation*}
\begin{aligned}
W_{R, K} = & \E \Big[ \sum_{j=0}^K  q( j | x, y_{0:K}; \theta^*)  \\
& \times \gtheta  \ssf{x}{y_0}{\theta^*} \gtheta \ssf{x}{y_j}{\theta^*}^\top\Big] \\
& =  \E \Big[ \sum_{j=0}^K  q( j | x, y_{0:K}; \theta^*) \\ 
& \times \gtheta \ssf{x}{y_j}{\theta^*}\gtheta  \ssf{x}{y_0}{\theta^*}^\top\Big]
\end{aligned}
\end{equation*}
where the expectation is w.r.t. $(x, y_0, \cdots, y_K)\sim p_{X, Y}(x, y_0)\prod_{j=1}^K p_N(y_j)$.
\label{lem:W_k}
\end{lemma}

{\em Proof:}
This follows directly from Lemma~\ref{lem:integral-rank}. $\qed$

\begin{lemma} 
\begin{equation*}
\begin{aligned}
 & \E [-\gtheta^2  L_R(\theta^*)] = \var [\gtheta L_R(\theta^*)] \\
 & =   \E [\gtheta \ssf{x}{y}{\theta^*}\gtheta \ssf{x}{y}{\theta^*}^\top] - W_{R, K}
\end{aligned}
\end{equation*} 
\label{lem:variance-rank}
\end{lemma}

{\em Proof:}
Define the shorthand, 
\[
q_j =  q( j | x, y_{0:K}; \theta^*).
\]
Notice that, 
\begin{equation*}
\begin{aligned}
 \gtheta L_R(\theta^*) & = \gtheta  \ssf{x}{y_0}{\theta^*} - \\
 &  \sum_{j=0}^K q_j \gtheta \ssf{x}{y_j}{\theta^*}.
\label{eq:first-derivative-rank}
\end{aligned}
\end{equation*}
We then have   
\begin{equation}
\begin{aligned}
   \gtheta^2 & L_R (\theta^*)  = \gtheta^2 \ssf{x}{y_0}{\theta^*}  \\
  & - \sum_{j=0}^K q_j \gtheta^2 \ssf{x}{y_j}{\theta^*} \\
  & - \sum_{j=0}^K q_j \gtheta \ssf{x}{y_j}{\theta^*}\gtheta \ssf{x}{y_j}{\theta^*}^\top\\
  & + \left(\sum_{j=0}^K q_j \gtheta \ssf{x}{y_j}{\theta^*}\right) \times \\ 
  & \left(\sum_{j=0}^K q_j \gtheta \ssf{x}{y_j}{\theta^*}\right)^\top \\
  & := I_1 - I_2 - I_3 + I_4
\end{aligned}
\nonumber
\end{equation}
By Lemma~\ref{lem:integral-rank}, 
\[
\E[I_1 - I_2] = 0
\]
\[
\E[I_3] = \E[\gtheta  \ssf{x}{y}{\theta^*} \ssf{x}{y}{\theta^*}^\top].
\]
Hence,
\begin{equation*}
\begin{aligned}
    & \E[-\gtheta^2 L_R (\theta^*)]  \\
 =  & \E [\gtheta \ssf{x}{y}{\theta^*}\gtheta \ssf{x}{y}{\theta^*}^\top] - W_{R, K}. 
\end{aligned}
\end{equation*}
By Lemma~\ref{lem:integral-rank}, $\E [L_R(\theta^*)] = 0$. Therefore, 
\[
\var \left[\gtheta L_R(\theta^*) \right] = \E \left[\gtheta L_R(\theta^*) \gtheta L_R(\theta^*)^\top\right].
\]
Observe that, 
\begin{equation}
\begin{aligned}
  & \gtheta L_R(\theta^*) \gtheta L_R(\theta^*)^\top = \\ 
  & \gtheta \ssf{x}{y_0}{\theta^*}\gtheta \ssf{x}{y_0}{\theta^*}^\top \\
   & - \sum_{j=0}^K  q_j\gtheta \ssf{x}{y_0}{\theta^*}\gtheta \ssf{x}{y_j}{\theta^*}^\top \\
   & -  \sum_{j=0}^K  q_j\gtheta \ssf{x}{y_j}{\theta^*}\gtheta \ssf{x}{y_0}{\theta^*}^\top\\
   & +  \left(\sum_{j=0}^K q_j \gtheta \ssf{x}{y_j}{\theta^*}\right) \times \\
   & \left(\sum_{j=0}^K q_j \gtheta \ssf{x}{y_j}{\theta^*}\right)^\top. 
\end{aligned}
\nonumber
\end{equation}
By Lemma~\ref{lem:W_k}, 
\begin{equation*}
\begin{aligned}
     \var & \left[\gtheta L_R(\theta^*) \right]  = \E \left[\gtheta L_R(\theta^*) \gtheta L_R(\theta^*)^\top\right] \\
     &   = \E \left[\gtheta \ssf{x}{y}{\theta^*}\gtheta \ssf{x}{y}{\theta^*}^\top\right] - W_{R, K}. 
\end{aligned}
\end{equation*}

% \begin{lemma}
% Under Assumption~\ref{assump:smoothness}, there exists constant $C$ such that 
% \[
% \sup_{(x, y)\in\mathcal{X}\times \mathcal{Y}} \Big\{ |\ssf{x}{y}{\theta^*}|, \vnorm{\gtheta \ssf{x}{y}{\theta^*}}, \vnorm{\gtheta^2 \ssf{x}{y}{\theta^*}}\Big\} \leq C. 
% \]
%   \label{lem:boundedness}
%   \vspace{-0.4cm}
% \end{lemma}
% {\em Proof:} 
% This follows from the facts that continuous functions are bounded on a compact set and ${\cal X}, {\cal Y}$ are finite. 

 \begin{lemma} Under Assumption~\ref{assump:boundedness}, there exists constant $C$ such that
\begin{equation*}
\begin{aligned}
\sup_{(x, y)\in\mathcal{X}\times \mathcal{Y}} \Big\{ |\ssf{x}{y}{\theta^*}|, \vnorm{\gtheta \ssf{x}{y}{\theta^*}},  \\
\vnorm{\gtheta^2 \ssf{x}{y}{\theta^*}}\Big\} \leq C.  
\end{aligned}
\end{equation*}
 \label{lem:boundedness}
 \end{lemma}

{\em Proof: } By definition, 
\[
\ssf{x}{y}{\theta^*} = \gtheta \str{x}{y}{\theta^*} + \log p_N(y).
\]
Then it follows that,
\[
\gtheta \ssf{x}{y}{\theta^*} = \gtheta \str{x}{y}{\theta^*},
\]
\[
\gtheta^2 \ssf{x}{y}{\theta^*} = \gtheta^2 \str{x}{y}{\theta^*}. 
\]
Since $p_N(y) > 0$ for all $y\in {\cal Y}$ and ${\cal Y}$ is finite, there exists constant $c$ such that
\[
p_N(y) > c.
\]
Therefore, 
\begin{equation*}
\begin{aligned}
& \sup_{(x, y)\in\mathcal{X}\times \mathcal{Y}}  |\ssf{x}{y}{\theta^*}| \\
\leq & \sup_{(x, y)\in\mathcal{X}\times \mathcal{Y}}  |\str{x}{y}{\theta^*}| + \log(1/c). 
\end{aligned}
\end{equation*}
Then the Lemma follows from Assumption~\ref{assump:boundedness}.

\begin{lemma}
  Under the assumptions in Theorem~\ref{thm:ranking-mse}, there exists an integer $K_0$ such that for some constant $C$ and all $K \geq K_0$, 
\begin{equation*}
\begin{aligned}
   & W_{R, K} = \E_X \Big[\left(\E_{Y|X = x}\gtheta \ssf{x}{y}{\theta^*}\right) \times  \\
   &  \left(\E_{Y|X = x}\gtheta \ssf{x}{y}{\theta^*}\right)^\top \Big] + \Delta_K
\end{aligned}
\end{equation*}
and 
\[
\|\Delta_K\| \leq C/\sqrt{K} 
\]
  \label{lem:W_rk_bound}
\end{lemma}
{\em Proof: } Define the abbreviation
\[
q_j =  q( j | x, y_{0:K}; \theta^*)
\]
and decompose $W_{R, K}$ into two terms, 
\begin{equation}
  \begin{aligned}
  & W_{R, K} \\
  = & \E\left[ \sum_{j=0}^K  q_j \gtheta  \ssf{x}{y_0}{\theta^*}\gtheta \ssf{x}{y_j}{\theta^*}^\top\right]\\
  = & \E\left[   q_0 \gtheta  \ssf{x}{y_0}{\theta^*}\gtheta  \ssf{x}{y_0}{\theta^*}^\top\right] \\
  + & \E\left[ \sum_{j=1}^K  q_j \gtheta  \ssf{x}{y_0}{\theta^*}\gtheta \ssf{x}{y_j}{\theta^*}^\top\right]\\
  & := J_1 + J_2.
  \end{aligned}
  \nonumber
  \end{equation}
By Lemma~\ref{lem:boundedness}, 
\begin{equation*}
\begin{aligned}
   \vnorm{J_1} &\leq  \sup_{x, y} \vnorm{ \gtheta  \ssf{x}{y}{\theta^*}\gtheta  \ssf{x}{y}{\theta^*}^\top}  \\
   & \times  \sup_{x, y} q_0\\
   & \leq C^2 \times \frac{\exp(C)}{K\exp(-C)} \leq \frac{C_1}{K}
\end{aligned}
\end{equation*}
% \[
%   \vnorm{J_1} \leq  \sup_{x, y} \vnorm{ \gtheta  \ssf{x}{y}{\theta^*}\gtheta  \ssf{x}{y}{\theta^*}^\top} \sup_{x, y} q(0| x, y_{0:K}; \theta^*) \leq C/K
% \]
By symmetry, 
\[
J_2 = K \E\left[ q_1 \gtheta  \ssf{x}{y_0}{\theta^*}\gtheta   \ssf{x}{y_1}{\theta^*}^\top\right]
\]
Substitute in the definition of $q_1$, 
\begin{equation*}
\begin{aligned}
 J_2 & = \sum_{x, y_{0:K}} p_{X}(x)p_{Y|X}(y_0|x)\prod_{i=1}^K p_N(y_i)  \\
 & \times \frac{\exp(  \ssf{x}{y_1}{\theta^*}) }{ \frac{1}{K}\sum_{j=0}^K \exp(\ssf{x}{y_j}{\theta^*} ) }  \\
 & \times \gtheta \ssf{x}{y_0}{\theta^*}\gtheta   \ssf{x}{y_1}{\theta^*}^\top  
\end{aligned}
\end{equation*}
Notice that 
\begin{equation*}
\begin{aligned}
\exp(  \ssf{x}{y_1}{\theta^*}) & = \exp(  \str{x}{y_1}{\theta^*} ) / p_N(y_1) \\
& = p_{Y|X}(y_1|x)Z(x; \theta^*) / p_N(y_1).
\end{aligned}
\end{equation*}
Then we can derive
    \begin{equation}
  \begin{aligned}
  & J_2 = \sum_{x, y_{0:K}} p_X(x)p_{Y|X}(y_0|x)p_{Y|X}(y_1|x)  \\
  &  \times\prod_{i=2}^K p_N(y_i)  \gtheta \ssf{x}{y_0}{\theta^*}\gtheta   \ssf{x}{y_1}{\theta^*}^\top   \\
  &  \times \frac{Z(x; \theta^*)}{\frac{1}{K}\sum_{i=0}^K \exp( \ssf{x}{y_i}{\theta^*})}  \\
  % & =  + \Delta
  % & = \sum_{x, y_0, y_1} \left\{ \frac{p_X(x)p_N(y_0)p_N(y_1) \exp( \ssf{x}{y_0}{\theta^*})\exp(  \ssf{x}{y_1}{\theta^*})}{Z(x; \theta^*)} ~ \times  \\
  % & \left. ~~~\gtheta \ssf{x}{y_0}{\theta^*}\gtheta   \ssf{x}{y_1}{\theta^*}^\top \times \frac{1}{Z(x; \theta^*)} \right\} + \Delta \\
  & := J_3 + \delta, 
  \end{aligned}
  \nonumber
  \end{equation} 
where 
\begin{equation*}
\begin{aligned}
& J_3  = \sum_{x, y_{0:K}} p_X(x)p_{Y|X}(y_0|x)p_{Y|X}(y_1|x)  \\
& \times \prod_{i=2}^K p_N(y_i) \gtheta \ssf{x}{y_0}{\theta^*}\gtheta   \ssf{x}{y_1}{\theta^*}^\top  \\
& = \sum_{x, y_0, y_1}  p_X(x)p_{Y|X}(y_0|x)p_{Y|X}(y_1|x) \\
&  \times \gtheta \ssf{x}{y_0}{\theta^*}\gtheta   \ssf{x}{y_1}{\theta^*}^\top   \\
& = \sum_{x} p_X(x)\left(\sum_{y}p_{Y|X}(y|x)\gtheta \ssf{x}{y}{\theta^*}\right)\\
&\times \left(\sum_{y}p_{Y|X}(y|x)\gtheta \ssf{x}{y}{\theta^*}\right)^\top\\
& = \E_X \left[ \left(\E_{Y|X = x}\gtheta \ssf{x}{y}{\theta^*}\right) \right.\\
& \left. \times  \left(\E_{Y|X = x}\gtheta \ssf{x}{y}{\theta^*}\right)^\top \right],
\end{aligned}
\end{equation*}
\begin{equation*}
\begin{aligned}
 & \delta = \sum_{ x, y_{0:K} } p_X(x)p_{Y|X}(y_0|x)p_{Y|X}(y_1|x)\prod_{i=2}^K p_N(y_i) \\
 & \times \gtheta \ssf{x}{y_0}{\theta^*}\gtheta   \ssf{x}{y_1}{\theta^*}^\top \\
 & \times \frac{Z(x; \theta^*) - \frac{1}{K}\sum_{i=0}^K \exp( \ssf{x}{y_i}{\theta^*})}{\frac{1}{K}\sum_{i=0}^K \exp( \ssf{x}{y_i}{\theta^*})} \\
 & = \sum_{x, y_{0:K}} p_X(x) p_{N}(y_0) p_{N}(y_1) \prod_{i=2}^K p_N(y_i)  \times \\ 
 & \times \frac{\exp(\ssf{x}{y_0}{\theta^*})}{Z(x; \theta^*)} \frac{\exp( \ssf{x}{y_1}{\theta^*})}{Z(x; \theta^*)}\\
 & \times  \frac{\gtheta \ssf{x}{y_0}{\theta^*}\gtheta   \ssf{x}{y_1}{\theta^*}^\top }{\frac{1}{K}\sum_{i=0}^K \exp( \ssf{x}{y_i}{\theta^*})} \\
 & \times \left(Z(x; \theta^*) - \frac{1}{K}\sum_{i=0}^K \exp( \ssf{x}{y_i}{\theta^*})\right).
\end{aligned}
\end{equation*}
By Lemma~\ref{lem:boundedness}, there exists constant $C_2$ such that
\begin{equation*}
\begin{aligned}
& \quad \sup_{x, y_{0:K}} \left\| \frac{\exp(\ssf{x}{y_0}{\theta^*})}{Z(x; \theta^*)}\frac{\exp( \ssf{x}{y_1}{\theta^*})}{Z(x; \theta^*)} \right. \\
& \left. \times \frac{\gtheta \ssf{x}{y_0}{\theta^*}\gtheta   \ssf{x}{y_1}{\theta^*}^\top }{\frac{1}{K}\sum_{i=0}^K \exp( \ssf{x}{y_i}{\theta^*})}\right\|  \\
& \leq  \frac{\exp(C)}{\exp(-C)} \times \frac{\exp(C)}{\exp(-C)}\times \frac{C^2}{\exp(-C)} \leq C_2.
\end{aligned}
\end{equation*}
We then have, 
\begin{equation*}
\begin{aligned}
 \left\| \delta\right\| & \leq C_2 \sum_{x, y_{0:K}} p_X(x) \prod_{i=0}^K p_N(y_i) \\
 & \times \Big|Z(x; \theta^*) - \frac{1}{K}\sum_{i=0}^K \exp \left(\ssf{x}{y_i}{\theta^*}\right) \Big | \\
 & \leq C_2 \E_{p_X(x)}\E_{\prod_{i=1}^K p_N(y_i)}  \Big|Z(x; \theta^*) -  \\
 &  \frac{1}{K} \sum_{i=1}^K \exp \left(\ssf{x}{y_i}{\theta^*}\right) \Big| + \frac{C_2}{K}\exp(C)\\
 & \leq C_2 \E_{p_X(x)} \Big(\E_{\prod_{i=1}^K p_N(y_i)} \Big|Z(x; \theta^*) - \\
 &  \frac{1}{K}\sum_{i=1}^K \exp( \ssf{x}{y_i}{\theta^*})\Big| ^2 \Big)^{1/2} + \frac{C_2\exp(C)}{K}.
\end{aligned}
\end{equation*}
Notice that for any given $x$ and $y_i\sim p_N(y_i)$, $\exp( \ssf{x}{y_i}{\theta^*}$ are I.I.D. random variables with expectation $Z(x; \theta^*)$. Therefore, by Lemma~\ref{lem:boundedness},
\begin{equation*}
\begin{aligned}
& \E_{\prod_{i=1}^K p_N(y_i)} \left|Z(x; \theta^*) - \frac{1}{K}\sum_{i=1}^K \exp( \ssf{x}{y_i}{\theta^*})\right| ^2  \\
& = \frac{1}{K}\var_{Y} \Big[\exp \left(\ssf{x}{y}{\theta^*}\right) \Big]  \\
&\leq \frac{1}{K}\E_{Y} \Big[\exp \left(2\ssf{x}{y}{\theta^*}\right) \Big]\\
& \leq \frac{\exp(2C)}{K}
\end{aligned}
\end{equation*}
and for some constant $C_3$, 
\[
 \left\| \delta\right\| \leq C_2\frac{\exp(C)}{\sqrt{K}} + \frac{C_2\exp(C)}{K+1}\leq \frac{C_3}{\sqrt{K}}
\]
Combining these facts together, 
\begin{equation*}
\begin{aligned}
& W_{R, K} = \E_X \Big[ \left(\E_{Y|X = x}\gtheta \ssf{x}{y}{\theta^*}\right)  \times \\
&  \left(\E_{Y|X = x}\gtheta \ssf{x}{y}{\theta^*}\right)^\top \Big] + \Delta_K 
\end{aligned}
\end{equation*}
where $\Delta_K = J_1 + \delta$ and for some constant $C_4$, 
\[
\|\Delta_K\| \leq \frac{C_4}{K}.
\]
$\qed$

\paragraph{Proof of Theorem~\ref{thm:ranking-mse}.}
By Lemma~\ref{lem:asym-rank} and \ref{lem:variance-rank}, 
\[
\sqrt{n}(\wh{\theta}_R - \theta^*)\rightarrow \mathcal{N} \left(0, \mathcal{I}_{R, K}^{-1} \right).
\]
where 
\[
\mathcal{I}_{R, K} = \E [\gtheta \ssf{x}{y}{\theta^*}\gtheta \ssf{x}{y}{\theta^*}^\top] - W_{R, K}.
\]
Notice that 

\begin{equation*}
\begin{aligned}
& \E \Big[\gtheta \ssf{x}{y}{\theta^*}\gtheta \ssf{x}{y}{\theta^*}^\top\Big] - \\
& ~~~~~~\E_X \Big[\left(\E_{Y|X = x}\gtheta \ssf{x}{y}{\theta^*}\right)  \\ 
& ~~~~~~~~~~~~\times \left(\E_{Y|X = x}\gtheta \ssf{x}{y}{\theta^*}\right)^\top \Big] \\
& = \E_X \E_{Y|X = x}  \Big[ \gtheta \ssf{x}{y}{\theta^*}\gtheta \ssf{x}{y}{\theta^*}^\top -  \\ 
&  ~~~~~~ \left(\E_{Y|X = x}\gtheta \ssf{x}{y}{\theta^*}\right)  \\ 
& ~~~~~~~~~~~~\times \left(\E_{Y|X = x}\gtheta \ssf{x}{y}{\theta^*}\right)^\top \Big ] \\
& = \E_X \Big[\var_{Y|X=x} \left[ \gtheta \ssf{x}{y}{\theta^*} \right] \Big] \\
& = \E_X \Big[\var_{Y|X=x} \left[ \gtheta \str{x}{y}{\theta^*} \right] \Big] \\
& = \fisher.
\end{aligned}
\end{equation*}
where the second last equation follows from the fact thet 
\[
\gtheta \str{x}{y}{\theta} = \gtheta \ssf{x}{y}{\theta}
\]
for all $x, y, \theta$. So by Lemma~\ref{lem:W_rk_bound}, 
\[
\mathcal{I}_{R, K} = \fisher - \Delta_K.
\]
Since $\fisher$ is non-singular, there exists positive integer $K_0$ such that for any $K\geq K_0$, $\sigma_{\min}(\fisher)\geq 2\|\Delta_K\|$. So $\mathcal{I}_{R, K}$ is positive definite and by Weyl's inequality, 
\[
\sigma_{\min}(\mathcal{I}_{R, K}) \geq \sigma_{\min}(\fisher) -  \vnorm{\Delta_K} \geq \sigma_{\min}(\fisher)/2. 
\]
Notice that,
\begin{equation*}
\begin{aligned}
\mathcal{I}_{R, K}^{-1} - \fisher^{-1}  & = \fisher^{-1} (\fisher - \mathcal{I}_{R, K}) \mathcal{I}_{R, K}^{-1} \\
&  = \fisher^{-1} \Delta_K \mathcal{I}_{R, K}^{-1}. 
\end{aligned}
\end{equation*}
Then we have 
\begin{equation*}
\begin{aligned}
& \vnorm{\mathcal{I}_{R, K}^{-1} - \fisher^{-1}} \leq \vnorm{\fisher^{-1}} \vnorm{\Delta_K} \vnorm{\mathcal{I}_{R, K}^{-1}}\\
& \leq  \frac{1}{\sigma_{\min}(\fisher)}\times \frac{C}{\sqrt{K}} \times \frac{2}{\sigma_{\min}(\fisher)}  \\
& \leq  \frac{2C}{c^2 \sqrt{K}}.
\end{aligned}
\end{equation*}
Finally, notice that $\wh{\theta}_R, \mle$ are asymptotically unbiased, 
\begin{equation*}
\begin{aligned}
  & | \mse_{\infty} (\wh{\theta}_R) -  \mse_{\infty} (\mle)|  \\
= & \tr(\mathcal{I}_{R, K}^{-1} - \fisher^{-1})/d  \\
\leq &  d \|\mathcal{I}_{R, K}^{-1} - \fisher^{-1}\|/d \\
\leq & \frac{2C}{c^2\sqrt{K}}
\end{aligned}
\end{equation*}

\subsection{Proofs for Binary Classification Loss}
\label{sec:proof-binary}

Recall that for any $x, y$ and parameter $(\theta, \gamma)$, we have defined 
\[
g(x, y; \theta, \gamma) = \frac{\exp(\ssf{x}{y}{\theta} - \gamma)}{\exp(\ssf{x}{y}{\theta} - \gamma) + K}.
\]

\begin{lemma}
Under Assumption~\ref{assump:binary}, 
\[
g(x, y; \theta^*, \gamma^*) 
= \frac{p_{Y|X}(y|x)}{p_{Y|X}(y|x) + K p_N(y)}
\]
\label{lemma:g}
\end{lemma}

{\em Proof:} Under Assumption~\ref{assump:binary}, 
\[
p_{Y|X}(y | x) = p_N(y) \exp(\ssf{x}{y}{\theta^*} - \gamma^*).
\]
Hence
\begin{equation*}
\begin{aligned}
&\frac{p_{Y|X}(y|x)}{p_{Y|X}(y|x) + Kp_N(y)}\\ 
= & \frac{p_N(y) \exp(\ssf{x}{y}{\theta^*} - \gamma^*)}{p_N(y) \exp(\ssf{x}{y}{\theta^*} - \gamma^*) + Kp_N(y)} \\
= &\frac{\exp(\ssf{x}{y}{\theta^*} - \gamma^*)}{\exp(\ssf{x}{y}{\theta^*} - \gamma^*) + K} 
\end{aligned}
\end{equation*}
$\qed$

\begin{lemma}
Define 
\[
p(y | x; \theta, \gamma) = \exp \{ \str{x}{y}{\theta} - \gamma \}
\]
For any $(\theta_1, \gamma_1), (\theta_2, \gamma_2)$, the following three statements are equivalent: 
\begin{enumerate}[label={(\arabic*)}]
  \item  For all $x, y$,
  \[
  p(y | x; \theta_1, \gamma_1) = p(y | x; \theta_2, \gamma_2)
  \]
  \item  For all $x, y$,
  \[
  g(x, y ; \theta_1, \gamma_1) = g(x, y ; \theta_2, \gamma_2)
  \]
  \item  For all $x, y$,
  \[
  \str{x}{y}{\theta_1} - \gamma_1 = \str{x}{y}{\theta_2} - \gamma_2.
  \]
\end{enumerate}
\label{lem:iden-binary}
\end{lemma}

{\em Proof:} 
Since $p_Y(y) > 0$ for all $y\in\mathcal{Y}$,  
\[
g(x, y; \theta, \gamma) = \frac{  p(y | x; \theta, \gamma)}{p(y | x; \theta, \gamma) + Kp_N(y)},
\]
and hence $(1) \Rightarrow (2)$. Observe that,  
\[
\ssf{x}{y}{\theta} - \gamma = \log \left(\frac{g(x, y; \theta, \gamma) }{1 - g(x, y; \theta, \gamma)}\right) + \log(K)
\]
and by definition
\[
\str{x}{y}{\theta} = \ssf{x}{y}{\theta} + \log p_N(y)
\]
We then have $(2) \Rightarrow (3)$. Finally, $(3)\Rightarrow (1)$ is obtained by plugging in the definition of $g(x, y; \theta, \gamma)$.
$\qed$

\subsubsection{Proof of Theorem~\ref{thm:pop-consistency-binary}}
\label{sec:pf-pop-consis-binary}
Notice that $L_B^\infty\left(\theta, \gamma \right) = \sum_{x, y} f(\theta, \gamma; x, y)$, where
\begin{equation*}
\begin{aligned}
& f(\theta, \gamma; x, y) \\
 = & p_{X, Y}(x, y)\log\left(g(x, y; \theta, \gamma) \right)  \\
& + K p_X(x)p_N(y) \log\left(1- g(x, y; \theta, \gamma)\right) \\
= &  \Big(p_{X, Y}(x, y) + K p_X(x)p_N(y)\Big) \\
& \left\{  \frac{p_{Y|X}(y|x)}{p_{Y|X}(y|x) + K p_N(y)} \log\left(g(x, y; \theta, \gamma) \right)  + \right. \\
& \left.  \frac{K p_N(y)}{p_{Y|X}(y|x) + K p_N(y)} \log\left(1- g(x, y; \theta, \gamma)\right)\right\}.
\end{aligned}
\end{equation*}
By Lemma~\ref{lem:entropy} and \ref{lemma:g}, $(\theta^*, \gamma^*) \in \arg\max f(\theta, \gamma; x, y)$ for all $x, y$. 
Then for any $(\bar{\theta}, \bar{\gamma}) \in\arg\max L_B^\infty(\theta, \gamma)$, it must be the case that $(\bar{\theta}, \bar{\gamma})$ maximizes $f(\theta, \gamma; x, y)$ for all $x, y$ simultaneously. It follows from Lemma~\ref{lem:entropy} that 
\begin{equation*}
\begin{aligned}
        & \frac{\exp(\ssf{x}{y}{\bar{\theta}} - \bar{\gamma})}{\exp(\ssf{x}{y}{\bar{\theta}} - \bar{\gamma}) + K} \\
       =  & \frac{p_{Y|X}(y|x)}{p_{Y|X}(y|x) + K p_N(y)}  \\
       =  & \frac{\exp(\ssf{x}{y}{\theta^*} - \gamma^*)}{ \exp(\ssf{x}{y}{\theta^*} - \gamma^*) + K}. 
\end{aligned}
\end{equation*}
By Lemma~\ref{lem:iden-binary}, this is equivalent to 
\begin{equation}
  \ssf{x}{y}{\bar{\theta}} - \bar{\gamma} = \ssf{x}{y}{\theta^*} - \gamma^* 
\label{eq:theta-gamma}
\end{equation}
which implies, 
\begin{equation}
  \str{x}{y}{\bar{\theta}} - \bar{\gamma} = \str{x}{y}{\theta^*} - \gamma^* 
\label{eq:theta-gamma}
\end{equation}
and therefore
\[
p(y|x ; \bar{\theta}, \bar{\gamma}) = p(y|x; \theta^*, \gamma^*),
\]
which proves the first part of the theorem. Under Assumption~\ref{assump:identifiability-binary}, equation~\eqref{eq:theta-gamma} implies $(\bar{\theta}, \bar{\gamma}) = (\theta^*, \gamma^*)$, that is $\Omega_B^*$ is a singleton. On the other hand, if there exists constant $c$ and parameter $\bar{\theta}$ such that for all $x, y$
\[
\str{x}{y}{\bar{\theta}} - \str{x}{y}{\theta^*} = c. 
\]
We then have  
\begin{equation*}
\begin{aligned}
\bar{\gamma} & = \log \left(\sum_{x, y} p_X(x)\exp(\str{x}{y}{\theta^*} + c)\right) \\
 & = c + \gamma^* 
\end{aligned}
\end{equation*}
and  
\[
\str{x}{y}{\bar{\theta}} - \bar{\gamma} = \str{x}{y}{\theta^*} - \gamma^*. 
\]
It follows from Lemma~\ref{lem:entropy} and \ref{lem:iden-binary} that $(\bar{\theta}, \bar{\gamma}) \in\arg\max L_B^\infty(\theta, \gamma)$. So if $\Omega^*_B$ is a singleton, it must be the case that $\bar{\theta} = \theta^*, \bar{\gamma} = \gamma^*$. This proves the second part of the theorem.

\subsubsection{Proof of Theorem~\ref{thm:consistency-binary}}
\label{sec:pf-consis-binary}
The proof is very similar to the proof of Theorem~\ref{thm:consistency-rank} and we leave out the details.

\subsubsection{Proof of Theorem~\ref{thm:mle-mse}}
\label{sec:pf-mle-mse}
The proof is very similar to the proof of Theorem~\ref{thm:ranking-mse} and the proof of Theorem~\ref{thm:binary-mse}. We leave out the details.

\subsubsection{Proof of Theorem~\ref{thm:binary-mse}}
\label{sec:pf-binary-mse}
Define $\beta = (\theta, \gamma), ~\wh{\beta} = (\wh{\theta}_B, \wh{\gamma}_B), \beta^* = (\theta^*, \gamma^*)$ and 
\[
\sbc{x}{y}{\beta} = \ssf{x}{y}{\theta} - \gamma - \log K.
\] 
Then the objective function $\objb(\theta, \gamma)$ could be written as the average of $n$ I.I.D. random variables $L_B^{(i)}(\beta), 1\leq i\leq n$. 
\begin{small}
  \begin{equation}
\begin{aligned}
& \objb(\theta, \gamma) = \frac{1}{n}\sum_{i=1}^n \left\{ \log\left(\frac{\exp(\sbc{x^{(i)}}{y^{(i, 0)}}{\beta})}{1+\exp(\sbc{x^{(i)}}{y^{(i, 0)}}{\beta})}\right) \right. \\
& \left. + \sum_{j=1}^{K}\log\left(\frac{1}{1+\exp(\sbc{x^{(i)}}{y^{(i, j)}}{\beta})}\right)\right\} \\
&  = \frac{1}{n}\sum_{i=1}^n  \Big\{ \log \sigma(\wt{s}(x^{(i)}, y^{(i)}; \beta))  \\
& + \sum_{j=1}^K \log (1-\sigma(\sbc{x^{(i)}}{y^{(i, j)}}{\beta})) \Big\}\\
& := \frac{1}{n} \sum_{i=1}^n L_B^{(i)}(\beta).
\end{aligned}
\end{equation}
\end{small}
Define 
\begin{equation*}
\begin{aligned}
L_B(\beta) &= \log \sigma( \sbc{x}{y_0}{\beta} ) \\ 
& + \sum_{j=1}^K \log (1-\sigma( \sbc{x}{y_j}{\beta}), 
\end{aligned}
\end{equation*}
where $(x, y_{0:K})\sim p_{X, Y}(x, y_0)\prod_{j=1}^K p_N(y_j)$. Then we have 
\[
(x^{(i)}, y^{(i, 0:K)}) =_d (x, y_0, \cdots, y_K)
\]
\[
L^{(i)}_B(\beta) =_d L_B(\beta), 1\leq i \leq n,
\]
where $=_d$ represents equal in distribution. 

Throughout the proof, we will repeatedly use the following facts 
\begin{equation}
  \sigma'(x) = \sigma(x)(1-\sigma(x)),
  \label{eq:derivative-sigmoid}
\end{equation}
\begin{equation}
\begin{aligned}
   & p_{X, Y}(x, y)(1-\sigma( \sbc{x}{y}{\beta^*})) \\
   = & K\px(x) p_N(y)\sigma(\sbc{x}{y}{\beta^*}).
\end{aligned}
  \label{eq:decomp-classify}
\end{equation}
\begin{lemma}
For any function $f(\cdot, \cdot): \mathcal{X}\times\mathcal{Y} \rightarrow \mathbb{R}^m, m\in\mathbb{N}_+$, 
\begin{equation*}
\begin{aligned}
  & \E \Big[ \left(1-\sigma( \sbc{x}{y_0}{\beta^*})\right) f(x, y_0)  \\ 
  & - \sum_{j=1}^K\sigma( \sbc{x}{y_0}{\beta^*} ) f(x, y_j)\Big] = 0
\end{aligned}
\end{equation*}
where $(x, y_{0:K})\sim p_{X, Y}(x, y_0)\prod_{j=1}^K p_N(y_j)$.
\label{lem:integral-classify}
\end{lemma}
{\em Proof: }
By definition, 
\begin{equation*}
\begin{aligned}
 & \E\left[\sum_{j=1}^K\sigma( \sbc{x}{y_j}{\beta^*}) f(x, y_j) \right] \\
= &  K\sum_{x, y} \px(x)p_N(y)\sigma(\sbc{x}{y}{\beta^*}) f(x, y).
\end{aligned}
\end{equation*}
By equality~\eqref{eq:decomp-classify}, 
\begin{equation*}
\begin{aligned}
    & \E\left[\sum_{j=1}^K\sigma( \sbc{x}{y_j}{\beta^*}) f(x, y_j) \right] \\
  = & \sum_{x, y} p_{X, Y}(x, y) (1-\sigma(\sbc{x}{y}{\beta^*}))f(x, y)\\
  = & \E[(1-\sigma(\sbc{x}{y_0}{\beta^*})) f(x, y_0)]
\end{aligned}
\end{equation*}
$\qed$

\begin{lemma}
  Under the assumptions in Theorem~\ref{thm:binary-mse}, as $n\rightarrow\infty$
\[
  \sqrt{n}\left(\wh{\beta}_B - \beta^*\right) \rightarrow \mathcal{N}\left(0, V_B\right)
\]
where $V_B$ is defined as 
\begin{equation*}
\begin{aligned}
\E \left[\gbeta^2 L_B(\beta^*)\right]^{-1} \var \left[\gbeta L_B(\beta^*) \right] \E \left[\gbeta^2 L_B(\beta^*)\right]^{-1} 
\end{aligned}
\end{equation*}
\label{lem:asym-binary}
\end{lemma}
{\em Proof: } The proof follows the same arguments as the proof of Lemma~\ref{lem:asym-rank} and we leave out the details.\\ 
$\qed$

Define the following quantities, 
\begin{equation*}
\begin{aligned}
\mubxk := \E_{Y|X=x} \Big[ \left(1-\sigma(\sbc{x}{y}{\beta^*})\right)  \\ 
\times \gbeta\sbc{x}{y}{\beta^*}\Big],
\end{aligned}
\label{eq:beta-x-K}
\end{equation*}
\begin{equation*}
\begin{aligned}
  \wbk & := \E_{p_{X, Y}}\Big[(1-\sigma(\sbc{x}{y}{\beta^*})) \times \\ 
  & \gbeta\sbc{x}{y}{\beta^*}\gbeta\sbc{x}{y}{\beta^*}^\top\Big].
  \label{eq:W-beta-K} 
\end{aligned}
\end{equation*}

\begin{lemma}
  \[
  \E[-\gbeta^2 L_B (\beta^*)] = \wbk
  \]
  \[
  \var \left[\gbeta L_B(\beta^*) \right] = \wbk - \frac{K+1}{K}\E_{X}[\mubxk  (\mubxk)^\top ]
  \]
  \label{lem:variance-binary}
\end{lemma}
{\em Proof: }
By straightforward calculation,
\begin{equation*}
\begin{aligned}
  \gbeta L_B(\beta^*) & = \left( 1- \sigma(\sbc{x}{y_0}{\beta^*}) \right) \gbeta\sbc{x}{y_0}{\beta^*}  \\
 & - \sum_{j=1}^K \sigma(\sbc{x}{y_j}{\beta^*}) \gbeta\sbc{x}{y_j}{\beta^*}  
\end{aligned}
\end{equation*}
\begin{equation*}
\begin{aligned}
  \gbeta^2 & L_B(\beta^*)  = -\sigma(\sbc{x}{y_0}{\beta^*}) (1-\sigma(\sbc{x}{y_0}{\beta^*})) \\
  & \times \gbeta\sbc{x}{y_0}{\beta^*}\gbeta\sbc{x}{y_0}{\beta^*}^\top\\
  & + (1-\sigma(\sbc{x}{y_0}{\beta^*}))\gbeta^2\sbc{x}{y_0}{\beta^*} \\
  &  -  \sum_{j=1}^K \sigma(\sbc{x}{y_j}{\beta^*})\gbeta^2 \sbc{x}{y_j}{\beta^*}\\
  & - \sum_{j=1}^K \sigma(\sbc{x}{y_j}{\beta^*}) (1-\sigma(\sbc{x}{y_j}{\beta^*}))  \\
  & \times \gbeta \sbc{x}{y_j}{\beta^*}\gbeta \sbc{x}{y_j}{\beta^*}^\top\\
  & := -I_1+I_2-I_3-I_4
\end{aligned}
\end{equation*}
By Lemma~\ref{lem:integral-classify}, 
\[
\E[I_2 - I_3] = 0.
\]
Since $(x, y_0)\sim p_{X, Y}(x, y_0)$ and $(x, y_j)\sim p_X(x)p_N(y_j)$ for all $j\geq 1$, 
\begin{equation*}
\begin{aligned}
 \E[I_1] &= \sum_{x, y}p_{X, Y}(x, y) (1-\sigma(\sbc{x}{y}{\beta^*}))\\
 & \times \sigma(\sbc{x}{y}{\beta^*})  \gbeta\sbc{x}{y}{\beta^*}\gbeta\sbc{x}{y}{\beta^*}^\top\\
  \E[I_4] &= K \sum_{x, y}\px(x)p_N(y)(1-\sigma(\sbc{x}{y}{\beta^*})) \\
  & \times  \sigma(\sbc{x}{y}{\beta^*})\gbeta\sbc{x}{y}{\beta^*}\gbeta\sbc{x}{y}{\beta^*}^\top
\end{aligned}
\end{equation*}
By equality~\eqref{eq:decomp-classify}, 
\begin{equation*}
\begin{aligned}
\E[I_4] & = \sum_{x, y}p_{X, Y}(x, y) (1-\sigma(\sbc{x}{y}{\beta^*}))^2 \\
& \times \gbeta\sbc{x}{y}{\beta^*}\gbeta\sbc{x}{y}{\beta^*}^\top
\end{aligned}
\end{equation*}
Therefore, 
\begin{equation*}
\begin{aligned}
\E[-\gbeta^2 & L_B (\beta^*)] = \E \left[I_1 + I_4\right] \\
& = \sum_{x, y} p_{X, Y}(x, y) (1-\sigma(\sbc{x}{y}{\beta^*})) \\ 
& \times \gbeta\sbc{x}{y}{\beta^*}\gbeta\sbc{x}{y}{\beta^*}^\top\\
& = \wbk
\end{aligned}
\end{equation*}
By Lemma~\ref{lem:integral-classify}, $\E [L_B(\beta^*)] = 0 $ and hence 
\[
\var \left[L_B(\beta^*)\right] = \E [\gbeta L_B(\beta^*) \gbeta L_B(\beta^*)^\top].
\]
Observe that,
\[
  \gbeta L_B(\beta^*) \gbeta L_B(\beta^*)^\top = I_1 - I_2 - I_2^\top + I_3
\]
where 
\begin{equation*}
\begin{aligned}
  I_1 & = (1-\sigma(\sbc{x}{y_0}{\beta^*}))^2 \\
  & \times \gbeta\sbc{x}{y_0}{\beta^*}\gbeta\sbc{x}{y_0}{\beta^*}^\top, \\
  I_2 & = \sum_{j=1}^K \sigma(\sbc{x}{y_j}{\beta^*})(1-\sigma(\sbc{x}{y_0}{\beta^*}))\\
  & \times \gbeta \sbc{x}{y_j}{\beta^*}\gbeta\sbc{x}{y_0}{\beta^*}^\top, \\
  I_3 & = \sum_{j=1}^K\sum_{l=1}^K \sigma(\sbc{x}{y_j}{\beta^*}) \sigma(\sbc{x}{y_l}{\beta^*}) \\
  & \times \gbeta \sbc{x}{y_j}{\beta^*}\gbeta \sbc{x}{y_l}{\beta^*}^\top.
\end{aligned}
\end{equation*}
By definition, 
\begin{equation*}
\begin{aligned}
\E[I_1] &= \sum_{x, y} p_{X, Y}(x, y)(1-\sigma(\sbc{x}{y}{\beta^*}))^2 \\
& \times \gbeta\sbc{x}{y}{\beta^*}\gbeta\sbc{x}{y}{\beta^*}^\top.
\end{aligned}
\end{equation*}
Notice that $(x, y_0, y_j)\sim p_{X, Y}(x, y_0)p_N(y_j)$, 
\begin{equation*}
\begin{aligned}
 \E[I_2] &= K \sum_{x, y, \wt{y}} p_{X, Y}(x, y)p_N(\wt{y}) \\
& \times \sigma(\sbc{x}{\wt{y}}{\beta^*})(1-\sigma(\sbc{x}{y}{\beta^*})) \\
& \times \gbeta \sbc{x}{\wt{y}}{\beta^*}\gbeta\sbc{x}{y}{\beta^*}^\top
\end{aligned}
\end{equation*}
By equation~\eqref{eq:decomp-classify},
\begin{equation*}
\begin{aligned}
\E & [I_2] = \sum_{x, y, \wt{y}} p_X(x)p_{Y|X}(y|x)p_{Y|X}(\wt{y}|x) \\ 
& \times (1-\sigma(\sbc{x}{\wt{y}}{\beta^*}))(1-\sigma(\sbc{x}{y}{\beta^*})) \\
& \times \gbeta \sbc{x}{\wt{y}}{\beta^*}\gbeta\sbc{x}{y}{\beta^*}^\top\\
& = \sum_{x} p_X(x) \mubxk \mubxk^\top\\
% & \times \left(\sum_{y} p_{Y|X}(y|x)(1-\sigma(\sbc{x}{y}{\beta^*}))\gbeta \sbc{x}{y}{\beta^*}\right)^\top\\
& =  \E_X[\mubxk  (\mubxk)^\top ].
\end{aligned}
\end{equation*}
We decompose $I_3$ into two parts:
\begin{equation*}
\begin{aligned}
I_3 &= \sum_{j=1}^K \sigma^2(\wt{s}(x, y_j;\beta^*)) \\ 
& \times \gbeta \sbc{x}{y_j}{\beta^*}\gbeta \sbc{x}{y_j}{\beta^*}^\top \\
& + \sum_{1\leq j\neq l\leq k} \sigma(\sbc{x}{y_j}{\beta^*}) \sigma(\sbc{x}{y_l}{\beta^*})\\
& \times \gbeta \sbc{x}{y_j}{\beta^*}\gbeta \sbc{x}{y_l}{\beta^*}^\top\\
& = I_3^1 + I_3^2
\end{aligned}
\end{equation*}
Notice that $(x, y_j)\sim \px(x)p_N(y_j)$, then again by equation~\eqref{eq:decomp-classify},
\begin{equation*}
\begin{aligned}
\E & [I_3^1] = K \sum_{x,y}\px(x)p_N(y) \sigma^2(\sbc{x}{y}{\beta^*}) \\ 
& \times \gbeta\sbc{x}{y}{\beta^*}\gbeta\sbc{x}{y}{\beta^*}^\top\\
& = \sum_{x,y} p_{X, Y}(x, y) (1-\sigma(\sbc{x}{y}{\beta^*})) \\
& \times \sigma(\sbc{x}{y}{\beta^*})\gbeta\sbc{x}{y}{\beta^*}\gbeta\sbc{x}{y}{\beta^*}^\top,
\end{aligned}
\end{equation*}
\begin{equation*}
\begin{aligned}
\E[I_1 + I_3^1] &= \sum_{x,y} p_{X, Y}(x, y)(1-\sigma(\sbc{x}{y}{\beta^*})) \\
& \times \gbeta\sbc{x}{y}{\beta^*}\gbeta\sbc{x}{y}{\beta^*}^\top= \wbk
\end{aligned}
\end{equation*}
Also notice that $(x, y_j, y_l) \sim \px(x)p_N(y_j)p_N(y_l)$ for $1\leq l\neq j \leq K$, 
\begin{equation*}
\begin{aligned}
\E[I_3^2] &= (K^2 - K)\sum_{x,y,\wt{y}}\px(x)p_N(y)p_N(\wt{y}) \\ 
& \times \sigma(\sbc{x}{y}{\beta^*})\sigma(\sbc{x}{\wt{y}}{\beta^*})\\
& \times \gbeta\sbc{x}{y}{\beta^*}\gbeta \sbc{x}{\wt{y}}{\beta^*}^\top\\
&= (1-1/K)\sum_{x} p_X(x) \mubxk  (\mubxk)^\top \\
% &= (1-1/K)\sum_{x} p_X(x) \left(\sum_{y} p_{Y|X}(y|x)(1-\sigma(\sbc{x}{y}{\beta^*}))\gbeta\sbc{x}{y}{\beta^*}\right)\times \\
% &\left(\sum_{\wt{y}} p_{Y|X}(\wt{y}|x)(1-\sigma(\sbc{x}{\wt{y}}{\beta^*}))\gbeta \sbc{x}{\wt{y}}{\beta^*}\right)^\top\\
& = (1-1/K)\E_X[\mubxk  (\mubxk)^\top].
\end{aligned}
\end{equation*}
where we apply equation~\eqref{eq:decomp-classify} to obtain the second equality. Combining these quantities, 
\begin{equation*}
\begin{aligned}
 \var \left[\gbeta L_B (\beta^*) \right]  & = \wbk  -  \\
 &  (1 + 1 / K)\E_{X}[\mubxk  (\mubxk)^\top ]
\end{aligned}
\end{equation*}
$\qed$

Define the following quantities,
\[
\mu = \E[ \gtheta  \ssf{x}{y}{\theta^*} ], ~\wt{\mu} = \begin{bmatrix} \mu \\ -1\end{bmatrix}, 
\]
\[
\mubx = \begin{bmatrix} \E_{Y|X=x}[ \gtheta  \ssf{x}{y}{\theta^*}] \\ -1 \end{bmatrix}, ~\wb = \begin{bmatrix} W & -\mu \\ -\mu^\top & 1 \end{bmatrix}
\]
\[
W = \E \big[\gtheta \ssf{x}{y}{\theta^*} \gtheta \ssf{x}{y}{\theta^*}^\top \big], 
\]
\[
\vbk = \wbk^{-1} \Big(\wbk - \frac{K+1}{K} \E_X \left[\mubxk  (\mubxk)^\top \right] \Big) \wbk^{-1}
\]
\[
~~\vb =  \wb^{-1}\left(\wb -  \mub\mub^\top \right)\wb^{-1}
\]
\begin{lemma}
  Under Assumption~\ref{assump:binary}, the Fisher information matrix can be simplified to 
  \[
    \fisher = \var_{X,Y}[\gtheta \str{x}{y}{\theta^*}].
  \]
  \label{lem:fisher}
\end{lemma}
{\em Proof: }
Notice that 
\begin{equation*}
\begin{aligned}
1 &= \sum_{y\in\mathcal{Y}} p_{Y|X}(y|x) = \sum_{y\in\mathcal{Y}} \exp(\str{x}{y}{\theta^*} + \gamma^*)\\
1 &= \sum_{x\in\mathcal{X}}\sum_{y\in\mathcal{Y}} p_{X, Y}(x, y) \\
& = \sum_{x\in\mathcal{X}} \sum_{y\in\mathcal{Y}} p_X(x) \exp(\str{x}{y}{\theta^*} - \gamma^*). 
\end{aligned}
\end{equation*}
Hence,
\begin{equation*}
\begin{aligned}
\exp(\gamma^*) & = \sum_{y\in\mathcal{Y}} \exp(\str{x}{y}{\theta^*}) \\
& = \sum_{x\in\mathcal{X}} \sum_{y\in\mathcal{Y}} p_X(x) \exp(\str{x}{y}{\theta^*}). 
\end{aligned}
\end{equation*}
Take derivative w.r.t. $\theta$ on both sides,
\begin{equation*}
\begin{aligned}
  & \sum_{y\in\mathcal{Y}} \exp(\str{x}{y}{\theta^*})\gtheta \str{x}{y}{\theta^*} \\
= & \sum_{x\in\mathcal{X}} \sum_{y\in\mathcal{Y}} p_X(x) \exp(\str{x}{y}{\theta^*})\gtheta \str{x}{y}{\theta^*} 
\end{aligned}
\end{equation*}
which implies
\begin{equation*}
\begin{aligned}
  & \E_{Y|X = x}[\gtheta \str{x}{y}{\theta^*}] \\
= & \sum_{x\in\mathcal{X}} \sum_{y\in\mathcal{Y}} p_X(x)\exp( \str{x}{y}{\theta^*} - \gamma^*) \\
& \times \gtheta \str{x}{y}{\theta^*} \\
= & \E_{X, Y} \left[ \gtheta \str{x}{y}{\theta^*}\right]. 
\end{aligned}
\end{equation*}
So $\E_{Y|X = x}[\gtheta \str{x}{y}{\theta^*}]$ is independent of $x$ and 
\[
\var_X \left[\E_{Y|X = x}[\gtheta \str{x}{y}{\theta^*}] \right] = 0. 
\]
Finally, by the law of total variance, 
\begin{equation*}
\begin{aligned}
  & \var_{X, Y}[\gtheta \str{x}{y}{\theta^*}] \\
= & \E_X \left[ \var_{Y|X=x}[\gtheta \str{x}{y}{\theta^*}]\right] \\ 
& + \var_{X} \left[ \E_{Y|X = x}[\gtheta \str{x}{y}{\theta^*}]\right]  \\
= & \E_X \left[ \var_{Y|X=x}[\gtheta \str{x}{y}{\theta^*}]\right] + 0\\
= & \fisher 
\end{aligned}
\end{equation*}
$\qed$

\begin{lemma} 
Under the assumptions in Theorem~\ref{thm:binary-mse}, $W$ and $\wb$ are non-singular and
\[
 \vb  = \begin{bmatrix} \fisher ^{-1}& \frac{W^{-1}\mu}{1-\mu^{\top}W^{-1}\mu} \\ \frac{\mu^\top W^{-1}}{1-\mu^{\top}W^{-1}\mu}&  \frac{\mu^{\top}W^{-1}\fisher W^{-1}\mu}{(1-\mu^{\top}W^{-1}\mu)^2} \end{bmatrix}
\]
\label{lem:w-mu-limit}
\end{lemma}
{\em Proof: }
Notice that, 
\begin{equation*}
\begin{aligned}
 W - \mu\mu^\top & = \E[\gtheta \ssf{x}{y}{\theta^*}\gtheta \ssf{x}{y}{\theta^*}^\top]  \\
 & -\E[\gtheta \ssf{x}{y}{\theta^*}]\E[\gtheta \ssf{x}{y}{\theta^*}]^\top \\
  & = \var[\gtheta \ssf{x}{y}{\theta^*}]
\end{aligned}
\end{equation*}
By Lemma~\ref{lem:fisher}, 
\[
 W =  \fisher + \mu\mu^\top. 
\]
Since $\fisher$ is non-singular, $W$ is strictly positive definite and by Sherman–Morrison formula, 
\[
W^{-1} = \fisher^{-1} - \frac{\fisher^{-1}\mu\mu^\top\fisher^{-1}}{1+\mu^\top\fisher^{-1}\mu}
\]
and
\[
\mu^\top W^{-1} \mu = \frac{\mu^\top\fisher^{-1}\mu}{1+\mu^\top\fisher^{-1}\mu} > 0.
\]
Therefore, by block matrix inversion formula, $\wb$ is non-singular and 
\begin{equation*}
\begin{aligned}
 \wb^{-1} &= \begin{bmatrix} W & -\mu\\ -\mu^\top & 1\end{bmatrix}^{-1} \\
 & = \begin{bmatrix} \fisher^{-1}& \frac{W^{-1}\mu}{1-\mu^{\top}W^{-1}\mu} \\ \frac{\mu^\top W^{-1}}{1-\mu^{\top}W^{-1}\mu}&  \frac{1}{1-\mu^{\top}W^{-1}\mu} \end{bmatrix}.
\end{aligned}
\end{equation*}
Notice that, 
\begin{equation*}
\begin{aligned}
\wb - \mub\mub^\top = \begin{bmatrix} \fisher  & 0\\ 0& 0\end{bmatrix}. 
\end{aligned}
\end{equation*}
We then have,
\begin{equation*}
\begin{aligned}
  & \wb^{-1}\left(\wb -  \mub\mub^\top \right)\wb^{-1} \\
= & \begin{bmatrix} \fisher ^{-1}& \frac{W^{-1}\mu}{1-\mu^{\top}W^{-1}\mu} \\ \frac{\mu^\top W^{-1}}{1-\mu^{\top}W^{-1}\mu}&  \frac{\mu^{\top}W^{-1}\fisher W^{-1}\mu}{(1-\mu^{\top}W^{-1}\mu)^2} \end{bmatrix}.
\end{aligned}
\end{equation*}
$\qed$

\begin{lemma}
Under the assumptions in Theorem~\ref{thm:binary-mse}, there exists an integer $K_0$ and positive constant $C>0$ such that for all $K\geq K_0$, $\wbk$ is non-singular and 
\begin{equation*}
\begin{gathered}
\sup_x\vnorm{\mubxk - \mubx}, ~\vnorm{\wbk^{-1} - \wb^{-1}} \leq C/K  \\
\vnorm{\E_X[\mubxk (\mubxk)^\top] - \E_X \left[\mubx\mubx^\top\right]} \leq C/K\\
\vnorm{\E_X[\mubxk (\mubxk)^\top]}, ~\vnorm{\wbk^{-1}} \leq C
\end{gathered}
\end{equation*}
\label{lem:w-mu-bound}
\end{lemma}

{\em Proof: }By definition and Lemma~\ref{lem:boundedness},
\begin{equation*}
\begin{aligned}
\sigma(\sbc{x}{y}{\beta^*}) & = \frac{\exp(\ssf{x}{y}{\theta^*} + \gamma^*)}{K + \exp(\ssf{x}{y}{\theta^*} + \gamma^*)}  \\
& \leq \frac{\sup_{x,y}\exp(\ssf{x}{y}{\theta^*} + \gamma^*)}{K} \\
& \leq \frac{C_1}{K}. 
\end{aligned}
\end{equation*}
Notice that, 
  \begin{equation*}
  \begin{aligned}
     & \mubxk =  \E_{Y|X = x} \begin{bmatrix} \gtheta \ssf{x}{y}{\theta^*} \\ 1\end{bmatrix}  \\
     &  - \E_{Y|X = x} \sigma(\sbc{x}{y}{\beta^*}\begin{bmatrix} \gtheta \ssf{x}{y}{\theta^*} \\ 1\end{bmatrix}  \\
   = & \mubx  - \Delta_{\mubxk} 
  \end{aligned}
  \end{equation*}
  where $\Delta_{\mubxk} = $
\begin{equation*}
\begin{aligned}
  \E_{Y|X = x} \left[\sigma(\sbc{x}{y}{\beta^*})\begin{bmatrix} \gtheta \ssf{x}{y}{\theta^*} \\ 1\end{bmatrix}\right]
\end{aligned}
\end{equation*}
Again by Lemma~\ref{lem:boundedness},
\begin{equation*}
\begin{aligned}
  \vnorm{\Delta_{\mubxk}} & \leq \frac{C_1}{K} \sup_{x,y}\vnorm{\begin{bmatrix} \gtheta \ssf{x}{y}{\theta^*} \\ 1\end{bmatrix}} \\
  &  \leq \frac{C_1 (C+1)}{K}.  
\end{aligned}
\end{equation*}
Observe that $\E_X \left[\mubxk (\mubxk)^\top\right] = \E_X \left[\mubx\mubx^\top\right] + \Delta_{\mu}$, where 
\begin{equation*}
\begin{aligned}
 \Delta_{\mu}  &=  \E_X \left[\mubx\Delta_{\mubxk}^\top \right]  + \E_X \left[\Delta_{\mubxk}\mubx^\top  \right] \\
 & + \E_X \left[\Delta_{\mubxk}\Delta_{\mubxk}^\top\right]. 
\end{aligned}
\end{equation*}
Similarly, by Lemma~\ref{lem:boundedness}, 
$\sup_x\vnorm{\mubx} \leq C + 1$. Therefore, there exist constants $C_3, C_4$ such that
\begin{equation*}
\begin{aligned}
\vnorm{\Delta_{\mu}} & \leq 2 \sup_x\vnorm{\mubx} \sup_x\vnorm{\Delta_{\mubxk}} \\
& + \sup_x\vnorm{\Delta_{\mubxk}}^2 \\
& \leq \frac{C_3}{K}, 
\end{aligned}
\end{equation*}
\[
\vnorm{\E_X \left[\mubxk (\mubxk)^\top\right]} \leq \sup_x\vnorm{\mubx}^2  + \vnorm{\Delta_{\mu}} \leq C_4. 
\]
On the other hand, $ \wbk = \wb + \Delta_{\wbk} 
$, where 
\begin{equation*}
\begin{aligned}
\Delta_{\wbk} & =  \E \Big[\sigma(\sbc{x}{y}{\beta^*})  \\
& \times \begin{bmatrix} \gtheta \ssf{x}{y}{\theta^*} \\ 1\end{bmatrix} \begin{bmatrix} \gtheta \ssf{x}{y}{\theta^*} \\ 1\end{bmatrix}^\top\Big].
\end{aligned}
\end{equation*}
By the same reasoning,
\begin{equation*}
\begin{aligned}
 & \vnorm{ \Delta_{\wbk} } \\
 \leq & \frac{C_1}{K}\sup_{x,y} \vnorm{\begin{bmatrix} \gtheta \ssf{x}{y}{\theta^*} \\ 1\end{bmatrix}\begin{bmatrix} \gtheta \ssf{x}{y}{\theta^*} \\ 1\end{bmatrix}^\top} \\
 \leq & \frac{C_1(C+1)^2}{K}.
\end{aligned}
\end{equation*}
Let $\sigma_{\min}(\cdot)$ denote the smallest singular value a matrix. By Lemma~\ref{lem:w-mu-limit}, $\wb$ is non-singular. 
So there exists $K_0$ such that $\sigma_{\min}\left(\wb\right) \geq 2 \vnorm{ \Delta_{\wbk} }$. Therefore, for any $K\geq K_0$, by Weyl's inequality, 
\begin{equation*}
\begin{aligned}
\sigma_{\min}\left(\wbk\right) &\geq \sigma_{\min}\left(\wb\right) -  \vnorm{ \Delta_{\wbk} } \\
& \geq \sigma_{\min}\left(\wb\right) / 2.  
\end{aligned}
\end{equation*}
Then
\begin{equation*}
\begin{aligned}
\vnorm{\wbk^{-1}} &= \frac{1}{ \sigma_{\min}\left(\wbk\right)} \leq \frac{2}{\sigma_{\min}\left(\wb\right)}. 
\end{aligned}
\end{equation*}
This implies, 
\begin{equation*}
\begin{aligned}
& \vnorm{\wbk^{-1} - \wb^{-1}} \\
= &  \vnorm{\wb^{-1} (\wb - \wbk) \wbk^{-1} } \\
\leq &  \vnorm{W_K^{-1}} \vnorm{\wb - \wbk} \vnorm{\wb^{-1}} \\
\leq &  \frac{2}{\sigma_{\min}\left(\wb\right)} \times \vnorm{\Delta_{\wbk}} \times \frac{1}{\sigma_{\min}(\wb)} \\
\leq &  \frac{2C_1(C+1)^2}{\sigma_{\min}(\wb)^2 K}\\
\leq &  \frac{C_5}{K}
\end{aligned}
\end{equation*}
for some constant $C_5$. $\qed$

\paragraph{Proof of Theorem~\ref{thm:binary-mse}.} By Lemma~\ref{lem:asym-binary} and Lemma~\ref{lem:variance-binary}, 
\[
\sqrt{n}(\wh{\beta} - \beta^*)\rightarrow \mathcal{N}(0, \vbk),
\]
\begin{equation*}
\begin{aligned}
& \vbk  - \vb = \left(\wbk^{-1} - \wb^{-1}\right)  \\
& - \frac{1}{K}\wbk^{-1}\E_X \left[\mubxk (\mubxk)^\top\right]\wbk^{-1} \\
& + \left(\wbk^{-1} - \wb^{-1}\right)\E_X \left[\mubxk (\mubxk)^\top\right]\wbk^{-1}  \\
& + \wb^{-1}\Big( \E_X \left[\mubxk (\mubxk)^\top\right] - \E_X \left[\mubx\mubx^\top\right] \Big) \wbk^{-1}\\
& + \wb^{-1}\E_X \left[\mubx\mubx^\top\right] \left(\wbk^{-1} - \wb^{-1}\right)
\end{aligned}
\end{equation*}
Applying Lemma~\ref{lem:w-mu-bound}, there exist constants C and integer $K'$ such that for some constant $C$ and all $K \geq K'$.
\[
\vnorm{\vbk - \vb} \leq C/K.
\]
Recall $\theta\in\reals^d$. Define $V_K$ as the upper $d\times d$ block of $\vbk$. Since $\fisher^{-1}$ is the upper $d\times d$ block of $\vb$, for all $K \geq K'$,
\[
\vnorm{V_K - \fisher^{-1}} \leq \vnorm{\vbk - \vb} \leq C / K.
\]
Choose $K_0$ large enough such that $K_0 \geq K'$ and 
\[
\sigma_{\min}(\fisher^{-1}) \geq 2C/K_0.
\]
Then by Weyl's inequality, for all $K\geq K_0$
\begin{equation*}
\begin{aligned}
\sigma_{\min}(V_K) & \geq \sigma_{\min}(\fisher^{-1}) - \vnorm{V_K - \fisher^{-1}}   \\
& \geq \sigma_{\min}(\fisher^{-1}) - C/K  \geq \sigma_{\min}(\fisher^{-1})/2. 
\end{aligned}
\end{equation*}
Therefore, $V_K$ is non-singular and we can define ${\cal I}_{B, K} = V_K^{-1}$. Then for all $K\geq K_0$
\[
\sqrt{n}(\wh{\theta}_B - \theta^*)\rightarrow \mathcal{N}(0, {\cal I}_{B, K}^{-1}).
\]
\[
\vnorm{{\cal I}_{B, K}^{-1} - \fisher^{-1}} = \vnorm{V_K - \fisher^{-1}}  \leq C / K.
\]
Finally, notice that $\wh{\theta}_B, \mle$ are asymptotically unbiased, 
\begin{equation*}
\begin{aligned}
  & |\mse_{\infty} (\wh{\theta}_B) -  \mse_{\infty} (\mle)|  \\
= & |\tr(\mathcal{I}_{B, K}^{-1} - \fisher^{-1})/d| \\
\leq & d \|\mathcal{I}_{R, K}^{-1} - \fisher^{-1}\|/d \leq C/ K.
\end{aligned}
\end{equation*}

\end{document}